\documentclass[]{gtech}

\usepackage[T1]{fontenc}
\usepackage[utf8]{inputenc}
\usepackage{CJKutf8}

\usepackage{amssymb}
\usepackage{url}
\usepackage{float}
\usepackage{xspace}
\usepackage{multicol}

\usepackage[table,xcdraw]{xcolor}  
\usepackage{colortbl}
\usepackage{hyperref}              

\usepackage{multirow}
\usepackage{bigdelim}
\usepackage{longtable}
\usepackage{tabularray}
\usepackage{wrapfig}
\usepackage{booktabs}
\usepackage{tabularx}
\usepackage{array}
\usepackage{makecell}
\usepackage{tabulary}
\usepackage{arydshln}

\usepackage[most]{tcolorbox} 
\usepackage[normalem]{ulem}
\useunder{\uline}{\ul}{}
\usepackage{todonotes}
\usepackage{fontawesome5}
\usepackage{bbding}
\usepackage{pifont}
\definecolor{headgray}{gray}{0.86}
\definecolor{groupgray}{gray}{0.94}
\definecolor{cotgray}{gray}{0.45}

\renewcommand{\title}[1]{\newcommand{\titlelist}{{\huge\fontfamily{optimistic}\selectfont #1}}}

\newcommand{\model}{PolyOCR\xspace}
\newcommand{\modelsmall}{\model-2B\xspace}
\newcommand{\modelbig}{\model-9B\xspace}

\definecolor{prompt}{HTML}{5f84e4}
\definecolor{img}{HTML}{820100}

\definecolor{CQColor}{rgb}{0.0,0.0,1.0}

\definecolor{TSColor}{rgb}{0.5,0.0,0.8}

\definecolor{CQRColor}{rgb}{1.0,0.0,1.0}

\newlength\savewidth

\newcommand{\modelicon}[1]{%
  \makebox[1.35em][c]{%
    \raisebox{-0.16\height}{\includegraphics[height=1.05em,width=1.05em,keepaspectratio]{#1}}%
  }\hspace{0.25em}%
}
\newcommand{\ourmodelicon}{%
  \makebox[1.35em][c]{%
    \raisebox{-0.22\height}{\includegraphics[height=1.53em,width=1.23em,keepaspectratio]{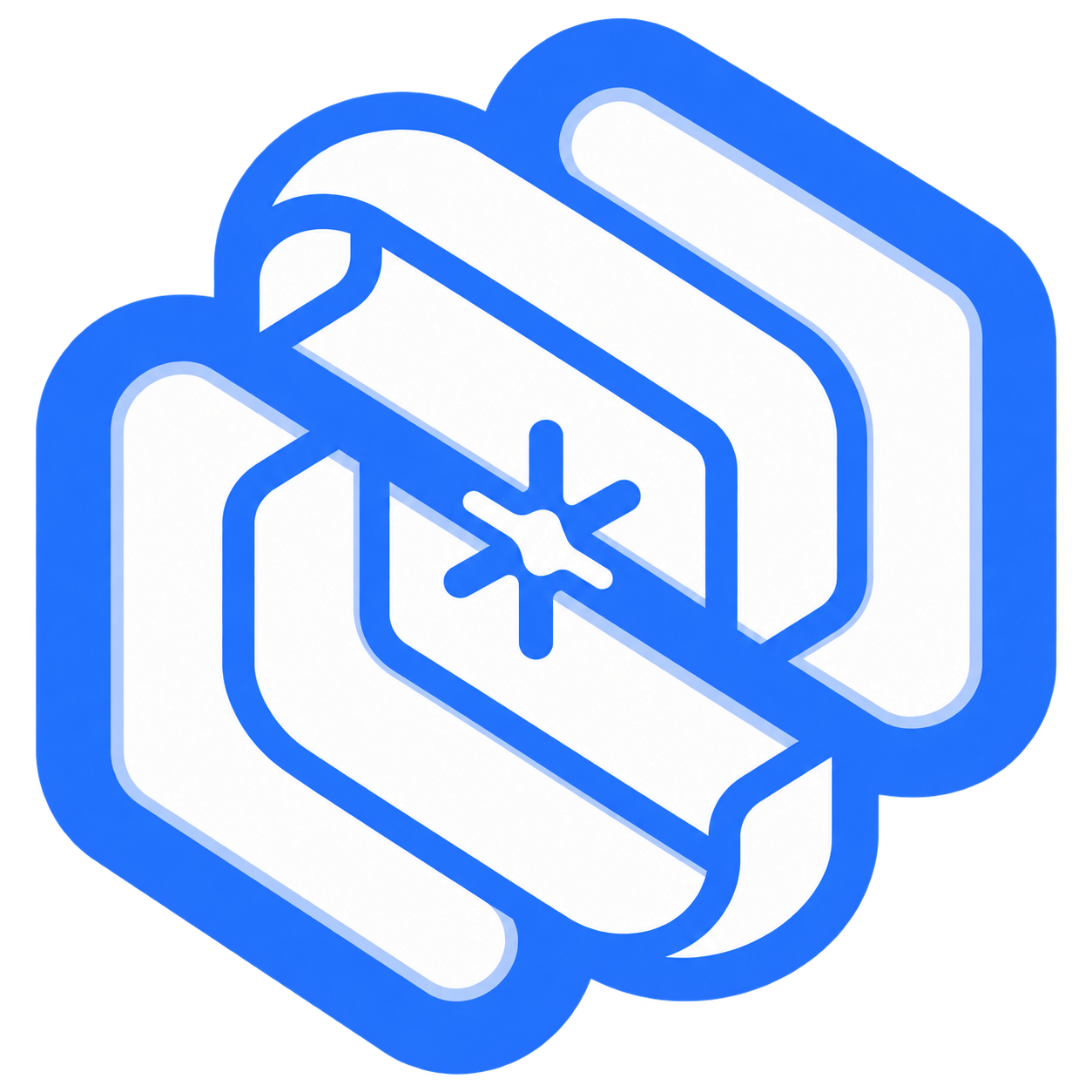}}%
  }\hspace{0.25em}%
}
\newcommand{\paddleicon}{%
  \makebox[1.35em][c]{%
    \raisebox{-0.04\height}{\includegraphics[width=1.22em,height=0.72em,keepaspectratio]{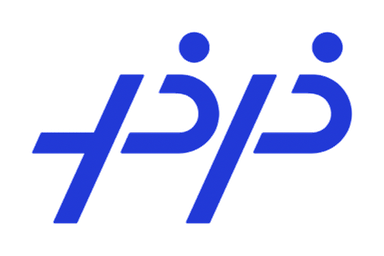}}%
  }\hspace{0.25em}%
}

\title{
\raisebox{-0.5em}{\includegraphics[height=1.5em]{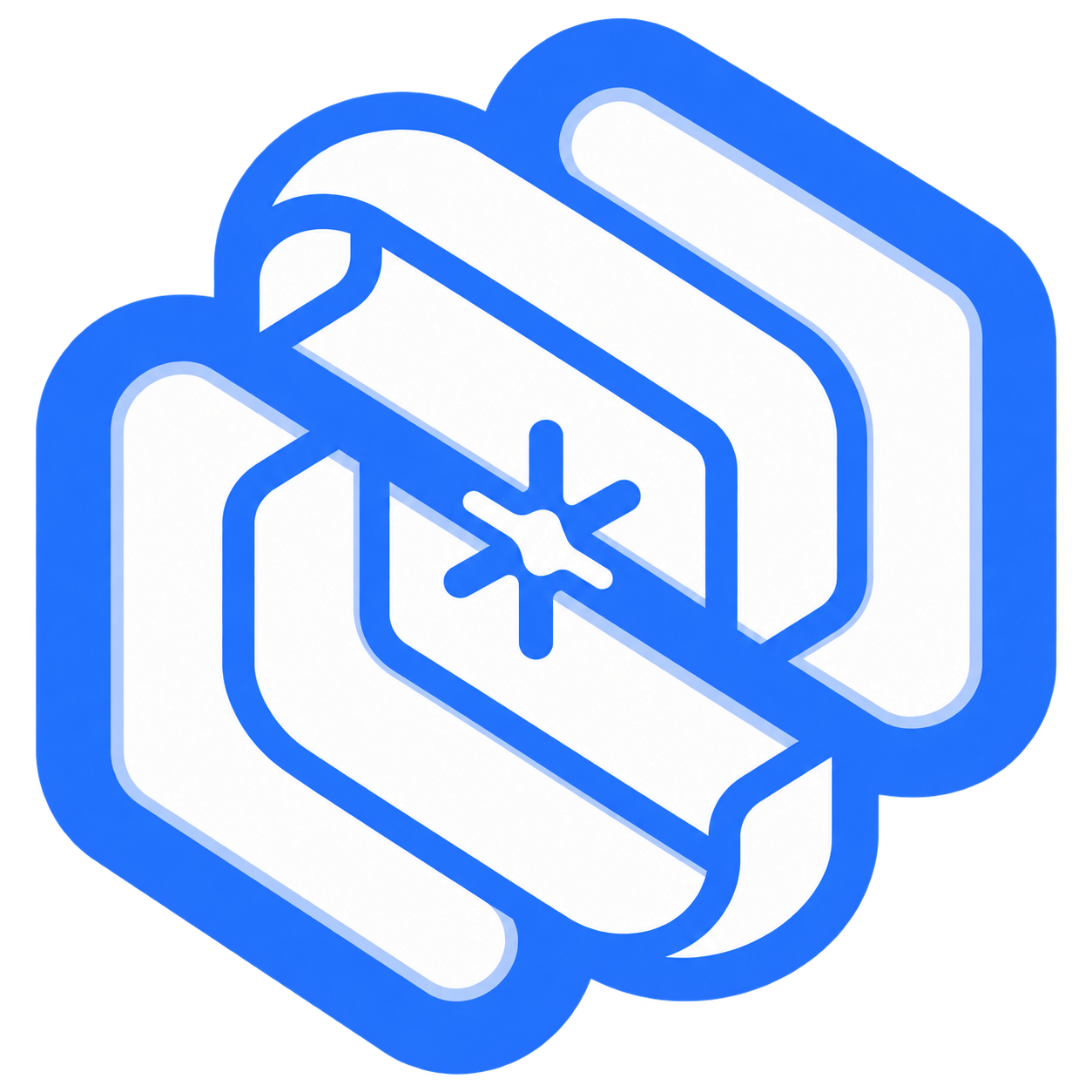}}\hspace{0.1em}
\textcolor[HTML]{0369ff}{\model-Venus}: Uni{f}ied OCR Foundation Models for Text-Centric Visual Intelligence
}

\author{GuangJian Team, ~Ant Group \vspace{-1.5em}}

\abstract{\fontsize{11pt}{12pt} \textit{Optical Character Recognition (OCR) is evolving from plain-text transcription toward general visual intelligence, requiring models to recognize, localize, and reason over textual information in complex visual environments. However, existing OCR systems often excel at only some tasks and struggle to balance recognition, parsing, and reasoning across scenarios.
In this report, we present \model, a family of unified OCR foundation models of varying scales. \model~combines a shared instruction-following framework with a large-scale data engine that converts heterogeneous visual resources into quality-verified OCR supervision.
We introduce \textit{Competence-Guided Policy Optimization}, which combines verifier-based Group Relative Policy Optimization with on-policy distillation through sample-wise routing based on teacher reliability and the teacher--student competence gap.
We also introduce OCRBench v2.1, our revision of OCRBench v2 with manually verified annotation corrections and task-aligned scoring metrics.
Extensive experiments across OCRBench v2.1, CC-OCR, in-house KIE Benchmark, OmniDocBench v1.6 and MDPBench demonstrate that \model achieves state-of-the-art or highly competitive performance.
}\vspace{-2mm}}

\date{September 30, 2026\vspace{-0.5mm}}

\gtechdata[Github]{%
    \faGithub \ \href{https://github.com/inclusionAI/PolyOCR-Venus}{\url{https://github.com/inclusionAI/PolyOCR-Venus}}%
\vspace{-0.5mm}}

\begin{document}
\maketitle

\vspace{-2mm}
\begin{figure}[H]
    \centering
    \includegraphics[width=\textwidth]{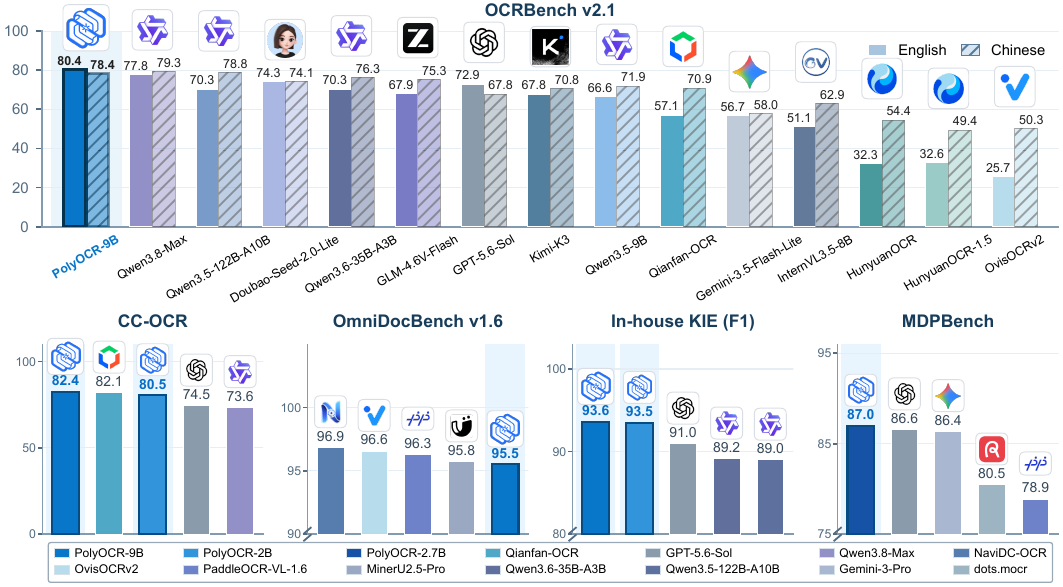}
    \vspace{-3mm}
    \caption{
    Cross-benchmark comparison. OCRBench v2.1 is shown above; CC-OCR, OmniDocBench v1.6, in-house KIE, and MDPBench are shown below with representative baselines. OmniDocBench uses the \model{}-9B layout-guided pipeline, whereas MDPBench uses \model{}-2.7B. CC-OCR excludes multilingual OCR; truncated axes are marked.
    }
    \label{fig:teaser}
\end{figure}

\section{Introduction}
\label{sec:intro}


Optical Character Recognition (OCR)~\citep{ocr1, ocr2} has evolved from conventional text transcription toward text-centric visual intelligence~\citep{qu2023towards, chen2024dtsm}. Traditional OCR systems mainly focus on converting text-containing images into machine-readable representations through modular pipelines consisting of text detection and recognition~\citep{chaudhuri2016optical, wei2024general}. Recent advances in multimodal large language models (MLLMs) have substantially expanded OCR capabilities beyond isolated recognition, enabling scene-text understanding, spatial localization, document structure recovery, table and formula parsing, information extraction, and OCR-oriented reasoning~\citep{ocrbenchv2, team2025hunyuanocr, li2026hunyuanocr, qu2026textshield}. As OCR increasingly serves as an interface between visual documents and intelligent systems, future OCR models require not only accurate text recognition but also broad capability coverage and robust generalization across diverse text-centric scenarios.

Despite significant progress, existing OCR systems remain limited in achieving such unified capabilities. Current approaches mainly follow two paradigms. The first consists of specialized OCR systems optimized for specific domains or task families, such as scene text recognition, document parsing, table understanding, and structured information extraction~\citep{team2025hunyuanocr,cui2025paddleocr, zhang2026paddleocr, zhong2026ocrverse, wang2026mineru2}. Although these systems achieve strong performance within their target settings, their task formulations, architectures, supervision schemes, and output formats are often highly specialized, resulting in fragmented capability development and limited knowledge transfer across heterogeneous OCR scenarios. The second paradigm is represented by general-purpose MLLMs, which provide unified interfaces and strong semantic reasoning abilities. However, OCR capabilities in these models are typically acquired implicitly through broad multimodal pretraining rather than systematically optimized across fine-grained text-centric tasks~\citep{achiam2023gpt, bai2025qwen3vl, xu2026deepseek}. Consequently, they may exhibit uneven performance in precise text recognition, spatial grounding, dense document understanding, and structured content extraction. These limitations reveal a fundamental challenge in balancing capability breadth, task-specific accuracy, and cross-scenario generalization within a single OCR model.

Building a unified OCR foundation model also introduces substantial training challenges. OCR-related tasks differ significantly in visual complexity, output structure, supervision granularity, and optimization objectives. Text recognition requires precise sequence generation, localization requires spatially grounded prediction, structured extraction requires strict format consistency, and OCR-centric reasoning emphasizes semantic understanding. Directly combining these heterogeneous tasks with fixed sampling strategies or static optimization objectives may lead to imbalanced learning dynamics, where dominant tasks overwhelm underrepresented or difficult capabilities. Moreover, positive transfer and task interference can coexist during multi-task learning, while sequential adaptation may cause catastrophic forgetting of previously acquired competencies~\citep{standley2020tasks, yu2020gradient, zhu2024modeltailor, jiang2025tabdsr, xin2026hytrechybridtemporalawareattention, xin2026unimomo}. Therefore, an effective OCR foundation model requires not only a unified task formulation but also an adaptive training strategy that coordinates heterogeneous supervision and maintains balanced capability acquisition.

To address these challenges, we present \model, a family of unified OCR foundation models with 2B and 9B parameter variants built upon Qwen3.5. \model~reformulates heterogeneous OCR and OCR-centric understanding tasks within a shared instruction-following and autoregressive generation framework. Through this unified formulation, diverse outputs, including text transcriptions, spatial coordinates, structured records, document representations, tables, formulas, translations, and open-form answers, can be generated through a common interface, enabling knowledge sharing across text recognition, text localization, document parsing, information extraction, visual text understanding, and OCR reasoning.

To enable such broad capability coverage, we develop a large-scale OCR data engine that addresses the limited coverage and imbalance of existing public and internal OCR datasets. The data engine converts heterogeneous visual assets into quality-verified supervision through controllable synthesis and model-assisted annotation, with particular emphasis on underrepresented scenarios and complex OCR capabilities that are insufficiently covered by existing resources. We further introduce \textit{Competence-Guided Policy Optimization} (CGPO), a competence-aware post-training framework that integrates verifier-based Group Relative Policy Optimization with on-policy distillation. CGPO dynamically adjusts the distillation weight according to teacher reliability and the evolving teacher--student competence gap, enabling selective and efficient utilization of teacher supervision.

To further extend multilingual document parsing capabilities, we additionally introduce \model{}-2.7B, a specialized configuration combining a 1.2B-parameter ViT vision encoder with a 1.5B-parameter Qwen2.5 text decoder. Trained on a curated in-house corpus covering 21 language/script groups, this variant is optimized for robust performance across diverse writing systems and acquisition conditions.

Extensive experiments on  OCRBench v2.1, CC-OCR, in-house KIE Benchmark, OmniDocBench v1.6 and MDPBench demonstrate that \model~achieves state-of-the-art or highly competitive performance across diverse OCR scenarios. The model family exhibits strong and balanced capabilities spanning text recognition, text localization, document parsing, information extraction, visual text understanding, and OCR reasoning, establishing the effectiveness of the proposed unified OCR training framework. The main contributions of this work are summarized as follows:

\begin{itemize}

\item We introduce \model, a family of unified OCR foundation models that integrate text recognition, text localization, document parsing, information extraction, visual text understanding, and OCR reasoning within a unified generative framework. The family includes general-purpose variants and a specialized multilingual configuration (\model{}-2.7B) optimized for diverse writing systems.

\item We develop a large-scale OCR data engine that converts heterogeneous visual resources into quality-controlled multi-task supervision through task-oriented construction, synthesis, and verification. Furthermore, we contribute to the community with refined and comprehensive benchmarks, including  OCRBench v2.1, CC-OCR, in-house KIE Benchmark, OmniDocBench v1.6 and MDPBench, to facilitate rigorous evaluation across general, multilingual, and industrial OCR scenarios.

\item We propose \textit{Competence-Guided Policy Optimization} (CGPO), a competence-aware post-training framework that dynamically combines verifier-based policy optimization and on-policy distillation through sample-wise supervision routing, effectively balancing heterogeneous task learning and preventing catastrophic forgetting.



\end{itemize}

\section{Related Work}

\subsection{Specialized OCR Systems}

Optical Character Recognition has traditionally been developed through task-specific formulations, including text detection, recognition, text spotting, handwriting recognition, multilingual OCR, and mathematical expression recognition~\citep{ocr1,ocr2}. Early OCR systems commonly adopted modular pipelines, where text localization, visual recognition, and post-processing were optimized independently. Representative approaches such as EAST~\citep{zhou2017east} and CRNN~\citep{shi2017crnn} established effective paradigms for scene-text detection and recognition, while recent Transformer-based models, including TrOCR~\citep{li2023trocr} and PARSeq~\citep{bautista2022parseq}, further improved recognition accuracy through large-scale pretraining and sequence modeling.

Beyond text recognition, specialized OCR models have been developed for diverse scenarios~\citep{qu2025revisiting, qu2024textsleuth}, including handwriting recognition, multilingual text understanding, formula recognition, text spotting, and structured information extraction. These systems typically employ task-specific architectures, supervision formats, decoding mechanisms, and evaluation protocols tailored to individual objectives. Although such specialization enables strong performance within specific domains, it also leads to fragmented capability development, where knowledge learned from one OCR task is difficult to transfer to others with different visual characteristics, output structures, and reasoning requirements.

The increasing diversity of OCR applications calls for a unified modeling paradigm that learns heterogeneous OCR capabilities within a shared framework~\citep{qu2024towards, qu2026omni}.

\subsection{Document Intelligence Models}

Document intelligence extends OCR from isolated text recognition toward page-level understanding, including layout analysis, reading-order modeling, structured reconstruction, and semantic information extraction~\citep{qu2026detect}. Early approaches such as LayoutLM~\citep{xu2020layoutlm} and LayoutLMv3~\citep{huang2022layoutlmv3} jointly modeled textual, visual, and spatial information, enabling effective document-level representation learning.

Another line of research explores OCR-free document understanding, where document images are directly converted into structured outputs without explicit OCR modules. Donut~\citep{kim2022donut} introduced an end-to-end document understanding framework based on image-to-sequence generation, while Nougat~\citep{blecher2024nougat} focused on converting scientific documents into structured markup representations. Recent document parsing systems further improve full-page understanding through structure-aware modeling and large-scale multimodal training, including MonkeyOCR~\citep{li2025monkeyocr}, Dolphin~\citep{feng2025dolphin,feng2026dolphin}, MinerU2.5~\citep{niu2026mineru2,wang2026mineru2}, and PaddleOCR-VL~\citep{cui2025paddleocrvl,zhang2026paddleocr}.

Despite significant progress in document-level understanding, existing document intelligence models primarily optimize for page-centric scenarios. Their capabilities are often concentrated on layout analysis, structural reconstruction, and document semantic extraction, while broader OCR intelligence, including natural-scene text perception, multilingual recognition, region-conditioned understanding, and OCR-oriented reasoning across heterogeneous visual environments, remains insufficiently unified.

\subsection{OCR-oriented Vision-Language Foundation Models}

The emergence of multimodal large language models provides a unified generative paradigm for visual perception, language understanding, spatial reasoning, and structured output generation. General-purpose vision-language models, including Gemini~\citep{comanici2025gemini25}, Qwen3-VL~\citep{bai2025qwen3vl}, InternVL3.5~\citep{wang2025internvl35}, and Ovis2.5~\citep{lu2025ovis25}, demonstrate strong capabilities across a wide range of visual understanding tasks. By adopting instruction following as a common interface, these models can flexibly handle image understanding, visual question answering, grounding, and document-related tasks within a unified framework.

Despite their broad multimodal capabilities, OCR competence in general-purpose vision-language models is typically acquired as an auxiliary capability from large-scale multimodal pretraining rather than through systematic optimization over diverse text-centric tasks. As a result, these models often exhibit uneven performance across fine-grained character recognition, dense text perception, precise localization, structured information extraction, and document-level reasoning. Comprehensive OCR evaluations, including OCRBench~\citep{liu2024ocrbench} and OCRBench v2~\citep{ocrbenchv2}, reveal substantial performance variations among different OCR capability dimensions, indicating that a unified multimodal interface alone is insufficient to establish balanced OCR intelligence.

To address this limitation, recent studies have explored OCR-oriented vision-language models that explicitly optimize text-centric visual understanding. TextMonkey~\citep{liu2026textmonkey} improves high-resolution text perception and grounding through OCR-oriented visual modeling. GOT-OCR2.0~\citep{wei2024general} extends OCR beyond plain-text transcription by supporting recognition, formulas, tables, charts, and region-conditioned understanding within a unified generative framework. HunyuanOCR~\citep{team2025hunyuanocr,li2026hunyuanocr} integrates text spotting, document parsing, information extraction, translation, and OCR-oriented question answering into a single OCR model. OCRVerse~\citep{zhong2026ocrverse} and Qianfan-OCR~\citep{dong2026qianfan} further expand OCR vision-language modeling toward broader task coverage and structured visual-text understanding.

These OCR-oriented vision-language models represent an important transition from isolated OCR modules toward unified generative OCR systems. Nevertheless, existing approaches still differ in their primary optimization objectives, with some emphasizing recognition and grounding, some focusing on document parsing, and others targeting selected extraction or reasoning tasks. A comprehensive OCR foundation model requires further integration of heterogeneous OCR capabilities, balanced supervision across diverse tasks, and adaptive optimization strategies that can maintain consistent performance from low-level text perception to high-level visual-text reasoning.

\subsection{OCR Data Construction and Capability-aware Post-training}

The development of OCR foundation models relies on large-scale and diverse supervision. Existing approaches typically combine manually annotated datasets with web-collected documents, scanned books, natural scene images, synthetic text images, rendered tables and formulas, automatically parsed documents, and model-generated instruction data~\citep{cui2025paddleocr,wang2026mineru2,dong2026qianfan, jiang2025ivy, fakehr1}. Synthetic generation and model-assisted annotation have become important for expanding coverage of long-tail scenarios, complex layouts, and structured output formats. However, constructing large-scale OCR supervision remains challenging due to annotation cost, heterogeneous task requirements, and the difficulty of maintaining consistent quality across diverse sources.

Recent OCR-oriented models commonly adopt multi-task supervised fine-tuning to align pretrained multimodal models with heterogeneous OCR tasks and output formats~\citep{licognicred}. Knowledge distillation and reinforcement learning have further been explored to improve recognition accuracy, structural validity, and instruction-following ability~\citep{team2025hunyuanocr,li2026hunyuanocr,zhang2026paddleocr,zhong2026ocrverse}. Nevertheless, OCR tasks differ substantially in data availability, visual complexity, annotation reliability, and optimization difficulty. Fixed data sampling strategies or static loss weighting may lead to imbalanced capability acquisition, where frequently optimized tasks dominate training while challenging or underrepresented capabilities remain insufficiently learned.

Recent OCR training pipelines provide more direct precedents for outcome optimization and distillation. olmOCR~2 uses the pass rate of verifiable unit tests as a GRPO reward, assessing text fidelity, reading order, table relations, and formula rendering~\citep{poznanski2025olmocr2}. Infinity-Parser2 performs joint reinforcement learning over eight document and multimodal tasks with task-specific rewards for textual, spatial, and structural accuracy~\citep{huang2026infinityparser2}. OvisOCR2 transfers an RL-enhanced 4B teacher to a 0.8B student through on-policy distillation with a reverse-KL objective evaluated on the student's top-$k$ support~\citep{lu2026ovisocr2}.

Beyond OCR, generalized knowledge distillation (GKD) already combines reinforcement learning with on-policy teacher supervision~\citep{agarwal2024opd}. More recent methods make this supervision conditional on observed outcomes: SCOPE routes verified student trajectories to perplexity-weighted self-reinforcement or distillation~\citep{zheng2026scope}, whereas Reward-Gated OPD selects trajectories according to agreement between verifier feedback and the teacher--student likelihood gap~\citep{akhondzadeh2026rewardgated}. CGPO instead estimates competence from teacher and student generations scored against a shared reference, calibrates teacher quality and probability of superiority on held-out generations, and uses the resulting sample-level weight to modulate OPD alongside an always-active GRPO objective.

Adaptive optimization methods have been investigated in broader multi-task and model training scenarios~\citep{qu2026dino, qu2025webly}. TwSG~\citep{jiang2026think} proposes the TL-GRPO scheme to optimize model performance on text-intensive understanding tasks such as ChartQA. GradNorm~\citep{chen2018gradnorm} balances task losses according to learning dynamics, PCGrad~\citep{yu2020gradient} mitigates gradient conflicts, DoReMi~\citep{xie2023doremi} optimizes data mixtures through distribution reweighting, and Skill-it!~\citep{chen2023skill} adapts training schedules according to skill dependencies. However, these approaches mainly focus on task balancing, data allocation, or curriculum design, without explicitly modeling the evolving competence relationship between teacher and student models during OCR-specific post-training~\citep{yu2024unified}. Building unified OCR foundation models therefore requires both comprehensive supervision and capability-aware optimization strategies that can adaptively allocate learning signals according to model competence.

\makeatletter
\renewcommand{\paragraph}{%
  \@startsection{paragraph}{4}{0pt}%
    {1.5ex plus 0.5ex minus 0.1ex}
    {-1em}
    {\normalfont\normalsize\bfseries}
}
\makeatother

\section{Data Construction}
\label{sec:data}

A unified OCR foundation model requires large-scale and diverse supervision that covers the full spectrum of text-centric visual intelligence capabilities, ranging from fundamental text recognition and spatial localization to structured content recovery, information extraction, multilingual transformation, and OCR-oriented reasoning. However, existing OCR datasets are typically constructed for individual tasks, visual domains, or languages, resulting in fragmented capability coverage and imbalanced supervision across different OCR scenarios.

To address this challenge, we develop a comprehensive data construction framework consisting of two complementary components: full-spectrum OCR data collection and targeted capability-driven data construction. First, we curate large-scale public datasets and in-house resources to establish broad coverage across OCR tasks, visual environments, languages, and annotation formats. Second, for capabilities that remain underrepresented due to limited public resources or expensive annotation requirements, we develop a targeted data-construction engine that generates controllable synthetic and model-assisted supervision. Together, these two components provide a unified and scalable data foundation for training OCR foundation models.

\subsection{Full-Spectrum OCR Data Collection}
\label{sec:full_spectrum_collection}

\begin{figure}[H]
\centering
\includegraphics[width=\textwidth]{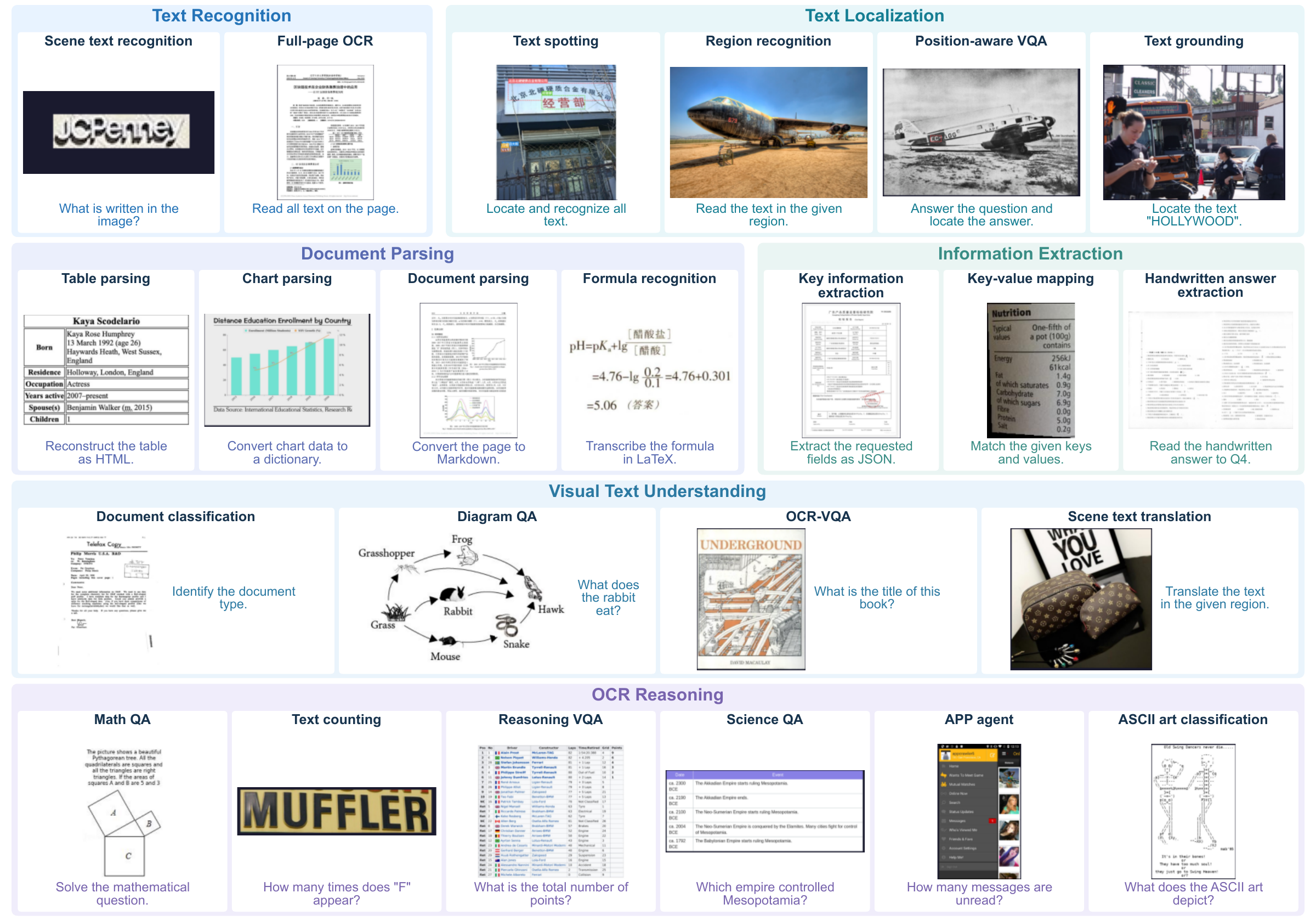}
\caption{Representative tasks across the six OCR capability dimensions. Each card pairs an example image with a shortened instruction. The gallery combines task illustrations from Table~\ref*{app:fine_grained_qa} with selected examples from OCRBench v2~\citep{ocrbenchv2}; the displayed examples illustrate task definitions rather than the composition of the training corpus.}
\label{fig:task_gallery}
\end{figure}

Inspired by Venus-DeFakerOne~\citep{DeFakerOne}, to establish broad capability coverage for unified OCR training, we curate OCR data from diverse visual domains and organize them according to six fundamental capability dimensions: text recognition, text localization, document parsing, information extraction, visual text understanding, and OCR reasoning. We use these six dimensions as a benchmark-independent organization of the training corpus. Figure~\ref{fig:task_gallery} illustrates representative tasks within these dimensions, and Table~\ref{tab:data_capabilities} summarizes their organization in the training corpus.

\begin{samepage}
The examples illustrate how a visual domain can support different tasks: a document may require transcription, structure recovery, field extraction, or question answering. Organizing data by the requested operation distinguishes these supervision requirements while retaining coverage across scenes, documents, charts, and application interfaces.
\end{samepage}

\paragraph{Text recognition.}
This category covers fundamental text perception capabilities across different visual conditions and granularity levels, including printed text recognition, scene text recognition, handwriting recognition, multilingual recognition, curved text recognition, and dense document transcription. The collected supervision spans character-, word-, line-, paragraph-, and page-level recognition tasks~\citep{ying2026beyond,karatzas2015icdar,chng2017totaltext,marti2002iam,liu2019ctw1500,union14m,yuan2022syntax}.

\paragraph{Text localization.}
This category focuses on spatially grounded text understanding, including text detection, text grounding, referring text localization, region-conditioned recognition, and text spotting. The corresponding annotations provide explicit spatial supervision represented by bounding boxes, quadrilaterals, or polygons, enabling joint learning of textual content and spatial information~\citep{karatzas2015icdar,chng2017totaltext,veit2016cocotext,long2022hiertext}.

\paragraph{Document parsing.}
This category includes document layout analysis, reading-order recovery, full-page document parsing, table reconstruction, chart parsing, and mathematical expression recognition. These tasks require models to recover hierarchical document structures beyond plain-text transcription, including both printed and handwritten mathematical content~\citep{zhong2019publaynet,zhong2020pubtabnet,smock2022pubtables,cheng2023m6doc,gervais2024mathwriting}.

\paragraph{Information extraction.}
This category covers structured information understanding from visually rich documents, including entity extraction, key information extraction, key--value association, form understanding, and semantic relation recovery from receipts, forms, certificates, and other structured document images~\citep{jaume2019funsd,huang2019sroie,xu2022xfund,kuang2023vie}.

\paragraph{Visual text understanding.}
This category extends OCR beyond perception toward higher-level visual-text interaction, including scene and document question answering, document classification, chart and table understanding, diagram interpretation, and visual text translation. These tasks require models to interpret textual information in the context of surrounding visual structures and perform task-oriented transformations~\citep{singh2019textvqa,biten2019stvqa,mathew2021docvqa,masry2022chartqa,mishra2019ocrvqa,liang2024dimt}.

\paragraph{OCR reasoning.}
This category targets reasoning capabilities grounded in visual text, including character and text-instance counting, arithmetic operations over visible values, spatial relationship reasoning, scientific question answering, and multi-step reasoning over scenes, documents, tables, charts, diagrams, and application interfaces~\citep{masry2022chartqa,lu2022scienceqa,lu2024mathvista,zhang2024mathverse}.

Although the collected corpus provides broad coverage across OCR tasks and visual scenarios, naturally available datasets remain highly imbalanced. In particular, capabilities such as spatially grounded bilingual translation, template-based key information extraction, Chinese document and table parsing, scene-text localization, and OCR-centric reasoning lack sufficient high-quality supervision due to limited public resources or expensive annotation processes. Therefore, we further develop a targeted data-construction engine to generate controllable synthetic and model-assisted training data for these underrepresented capabilities.

\begin{table*}[t]
\centering
\caption{Capability organization of the full-spectrum OCR training corpus.}
\label{tab:data_capabilities}
\renewcommand{\arraystretch}{1.15}
\setlength{\tabcolsep}{4pt} 
\fontsize{10pt}{11pt}\selectfont                   
\begin{tabularx}{\textwidth}{@{} l X @{}}
\hline
\textbf{Capability group} & \textbf{Representative tasks} \\
\hline
Text Recognition
& Scene text, handwriting, multilingual OCR, full-page transcription \\
Text Localization
& Detection, grounding, referring, region recognition, text spotting \\
Document Parsing
& Document, layout, table, formula, chart parsing \\
Information Extraction
& Entity extraction, KIE, key--value association, form understanding \\
Visual Text Understanding
& Document VQA, chart/table understanding, classification, translation \\
OCR Reasoning
& Counting, numerical calculation, spatial reasoning, multi-step inference \\
\hline
\end{tabularx}
\end{table*}

\subsection{Targeted Data-Construction Engine}
\label{sec:data_engine}

Although full-spectrum data collection provides broad OCR coverage, naturally available datasets remain insufficient for several capabilities due to limited annotations and high labeling costs. To address these gaps, we develop a targeted data-construction engine that generates capability-specific synthetic and model-assisted supervision. Each pipeline is designed to preserve the semantic requirements, spatial constraints, and output formats of its target OCR task.

\begin{figure}[!htbp]
\centering
\includegraphics[width=\textwidth]{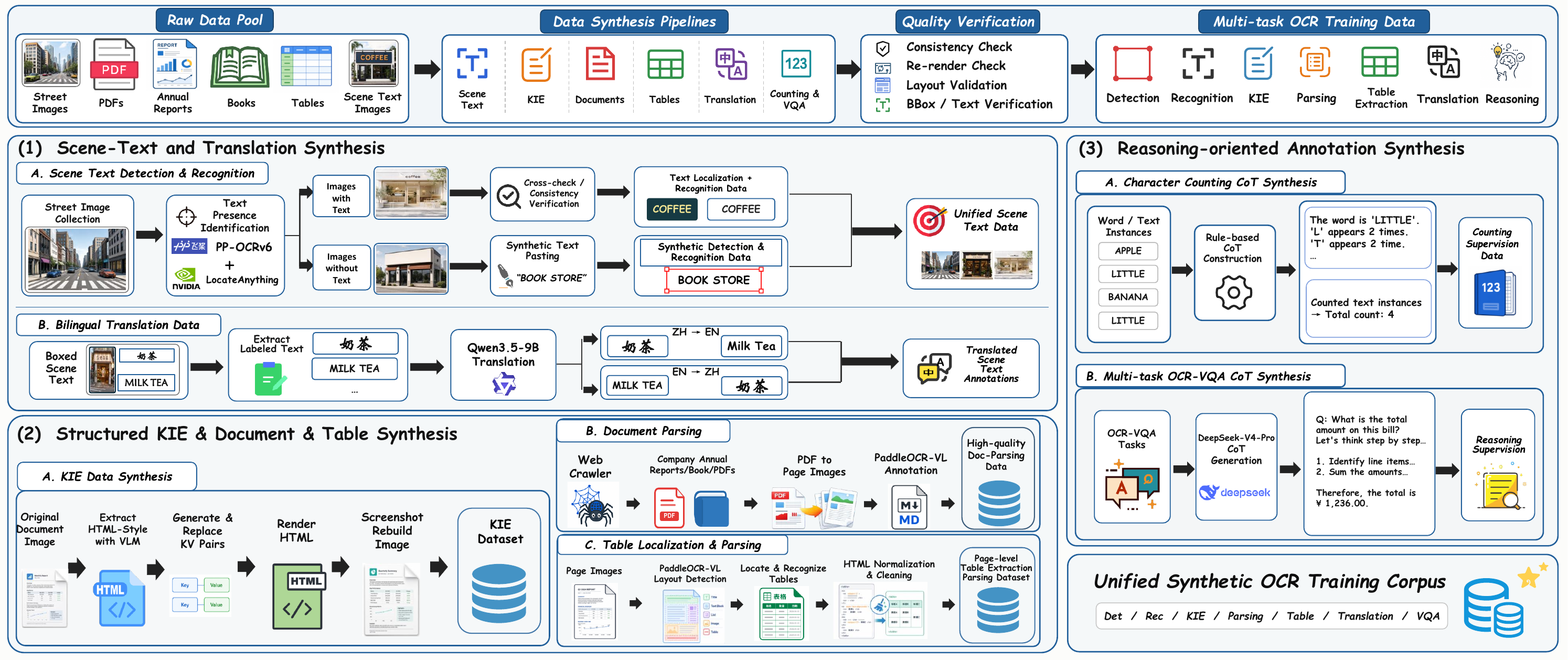}
\caption{Overview of the targeted data-construction engine. Heterogeneous raw assets are transformed into task-specific supervision for scene-text detection and recognition, bilingual translation, key information extraction, document parsing, Chinese table localization and parsing, and OCR-centric reasoning.}
\label{fig:data_engine}
\end{figure}

As illustrated in Figure~\ref{fig:data_engine}, the engine decouples raw visual resources from task-specific supervision generation. Diverse sources, including street images, document templates, annual reports, books, PDFs, tables, and verified OCR annotations, are transformed into supervision for scene-text understanding, bilingual translation, key information extraction, document parsing, table understanding, and OCR-centric reasoning. The engine is organized into three complementary synthesis pipelines: scene-text and translation synthesis, structured KIE, document, and table synthesis, and reasoning-oriented annotation synthesis. Each pipeline incorporates dedicated generation and verification procedures to ensure visual-text consistency, structural validity, and task-specific supervision quality.

\paragraph{(1) Scene-Text and Translation Synthesis}

\paragraph{A. Scene Text Detection \& Recognition.} Rendering text into natural backgrounds provides paired localization and recognition supervision, as demonstrated by SynthText~\citep{gupta2016synthtext}. We construct scene-text supervision from both real-world images and controllable synthesis. For text-containing natural images, PP-OCRv6~\citep{ppocrv6} and LocateAnything~\citep{wang2026locateanything} are employed as independent annotators for text localization and recognition. Their predictions are cross-validated through spatial overlap and transcription consistency, and only mutually verified instances are retained as localization, recognition, and spotting supervision. Samples exhibiting substantial disagreement are identified as hard cases and further annotated by human annotators before being incorporated into the training corpus. For images without visible text, a synthetic branch renders text onto realistic background regions while controlling language, font, character size, orientation, color, contrast, perspective distortion, blur, and compression artifacts. The rendered transcriptions and spatial annotations are further verified for visual-text consistency before being added to the training set. This dual-source construction strategy provides both naturally occurring scene-text diversity and controllable localization-recognition supervision.

\paragraph{B. Bilingual Translation Data.} We construct spatially grounded translation supervision from scene images with verified text regions and transcriptions. For each text instance, Qwen3.5-9B generates both Chinese--English and English--Chinese translations. The generated translations are filtered using language consistency, numerical and symbol preservation, named-entity consistency, repetition detection, and length constraints. Each sample contains the original image, a region-aware translation instruction, and the corresponding translation. This construction enables the model to jointly learn text localization, recognition, and multilingual transformation while maintaining the spatial grounding of the source text.

\paragraph{(2) Structured KIE \& Document \& Table Synthesis}

\paragraph{A. KIE Data Synthesis.} Recent KIE synthesis frameworks such as SAYRE use document exemplars and editable HTML representations to generate new images with aligned field annotations~\citep{xu2026sayre}. We develop a template-preserving synthesis pipeline for structured documents, including forms, receipts, certificates, labels, and other field-oriented images. A vision-language model first reconstructs each document into an editable HTML/CSS representation. A language model then replaces the original key--value contents while preserving field semantics and layout constraints. The modified templates are subsequently rendered into new document images with naturally aligned annotations. Generated samples are filtered for text overflow, abnormal line wrapping, element overlap, missing fonts, field inconsistency, and severe visual distortion. This pipeline expands the semantic and visual diversity of KIE supervision while preserving accurate key--value alignment.

\paragraph{B. Document Parsing.} 

To enrich document-level supervision, we collect a diverse corpus of PDF documents, including academic papers, annual reports, research reports, books, and documents containing handwritten content, and render their pages as images. To promote page diversity, we first extract page features using a visual encoder and apply k-means clustering to group pages with similar visual characteristics. In parallel, we use PP-DocLayoutV3 to detect layout elements and compute a page complexity score based on their counts and predefined category weights, assigning higher weights to complex, less frequent element types, such as figures and charts, and lower weights to common, simpler elements, such as regular text. Within each cluster, we stratify pages by their complexity scores and sample uniformly across strata, ensuring coverage of diverse visual characteristics and complexity levels.We then independently parse the sampled pages using PaddleOCR-VL and MinerU2.5-Pro to produce document annotations in reading order~\citep{cui2025paddleocr,cui2025paddleocrvl,wang2026mineru2}. We compare the outputs of the two models and use their similarity as a proxy for annotation quality. Annotations for pages with high inter-model agreement are retained as high-quality candidates. For pages with low agreement, we use Qwen-3.5 to identify and correct omissions, recognition errors, and structural discrepancies, taking both parsing outputs and the original page image as input. The resulting annotations are consolidated into Markdown and plain-text formats. Markdown annotations preserve document hierarchy and content elements, including paragraphs, headings, notes, lists, tables, and formulas, while plain-text annotations support full-page transcription tasks. All annotations undergo rule-based cleaning and normalization, including LaTeX standardization for formulas, HTML normalization for tables, removal of unsupported tags and placeholders, elimination of malformed structures, and deduplication.


\paragraph{C. Table Localization \& Parsing.} Existing datasets for table understanding primarily focus on table structure recognition, offering limited support for joint table localization and parsing within full-page documents. To address this limitation, we construct a full-page table understanding dataset covering diverse visual styles and levels of structural complexity. We first detect tables in document pages and classify them as fully bordered, borderless, or partially bordered using a table classification model. Each table is then independently parsed by PaddleOCR-VL and MinerU2.5-Pro~\citep{cui2025paddleocr,cui2025paddleocrvl,wang2026mineru2}, and their outputs are normalized into HTML. We compute the TEDS score between the corresponding outputs to assess inter-model consistency as a reference for annotation quality. Annotations with high consistency are accepted as candidates, while those with lower consistency are refined by Qwen-3.5 using the original page image and both models’ outputs. To preserve annotation completeness, we retain a page only after all its tables have been annotated and passed the quality checks, discarding pages with unresolved omissions or severe annotation errors. From the resulting HTML annotations, we extract row and column counts, the number and proportion of merged cells, and their row and column spans to characterize table size and structural complexity. We use these attributes together with the visual style categories for stratified sampling, followed by manual spot checks to promote data diversity and annotation quality. Each sample comprises a full-page document image, table bounding boxes, and HTML annotations, supporting joint table localization and structure parsing.


\paragraph{(3) Reasoning-oriented Annotation Synthesis}

\paragraph{A. Character Counting CoT Synthesis.} To provide reliable supervision for OCR-related quantitative reasoning, we construct counting tasks from quality-verified text recognition and localization annotations using deterministic templates. Text annotations are normalized with respect to character encoding, whitespace, punctuation, and full-width or half-width symbols while preserving character case and visual semantics. Based on these annotations, we automatically generate diverse counting instructions and reasoning traces, including character-level counting and text-instance counting. The generated samples inherit exact answers from verified OCR annotations, providing consistent supervision for OCR-centric numerical reasoning without introducing noise from model-generated labels.

\paragraph{B. Multi-task OCR-VQA CoT Synthesis.} For complex reasoning tasks that cannot be efficiently generated through deterministic rules, we employ strong reasoning models, represented by DeepSeek-V4-Pro~\citep{xu2026deepseek}, to construct candidate questions, explanations, and answers based on verified OCR and structured annotations. The generated examples cover textual content understanding, document field reasoning, key--value relationships, table interpretation, spatial reasoning, and other OCR-centric capabilities. The source annotations are used only for generation guidance and verification. Candidate samples are validated through task-specific consistency checks, and examples containing incorrect answers, unsupported reasoning steps, or invalid output formats are removed.

\subsection{Cross-Pipeline Quality Control}
\label{sec:data_quality_control}

Across the construction pipelines, quality control uses three complementary forms of verification. First, agreement-based validation compares independent annotations when multiple recognizers or parsers are available, retaining mutually consistent predictions and sending selected scene-text disagreements for human adjudication. Second, structure-aware validation checks whether generated coordinates, HTML, Markdown, formulas, and key--value records remain well formed and consistent with the rendered image. This includes detecting element overlap, text overflow, malformed table structures, missing fields, unsupported tags, and content loss. Third, source-grounded validation checks whether generated translations, questions, reasoning traces, and answers preserve the textual or structured evidence available in the source annotations. Samples that fail the corresponding checks are removed rather than repaired through unconstrained generation.

These validation signals also expose difficult examples. In particular, cross-model disagreements that are resolved through human annotation provide quality-verified hard scene-text cases, while valid samples with challenging structured outputs are retained only after passing task-specific format and consistency checks. During training, such hard examples are mixed with broader coverage data without being allowed to dominate the task distribution. After capability balancing and sampling, the resulting corpus delivers approximately 60 million high-quality OCR training instances, achieving broad coverage across diverse scenarios, tasks, and supervision formats to support model training.

\section{Training}
\label{sec:training}

\subsection{Training Pipeline}
\label{sec:training_pipeline}

Training a unified OCR foundation model is challenging due to the highly heterogeneous nature of OCR tasks, where different capabilities exhibit distinct learning difficulties, optimization objectives, and competence boundaries. A model may achieve strong performance on common recognition patterns while still struggling with complex document structures, long-tail scenarios, or reasoning-intensive OCR tasks.

To address this challenge, we develop a two-stage training framework consisting of capability acquisition and competence-guided refinement. The first stage adopts supervised fine-tuning (SFT) to establish broad OCR capabilities across heterogeneous tasks, including text recognition and localization, document parsing, table understanding, key information extraction, multilingual transformation, and OCR-centric visual question answering. The second stage introduces Competence-Guided Policy Optimization (CGPO), which combines verifier-based Group Relative Policy Optimization (GRPO) with adaptive on-policy distillation (OPD) to dynamically allocate optimization signals according to the competence relationship between the teacher and student models.

Both model variants use the same SFT and CGPO training settings. The CGPO stage uses a learning rate of $10^{-6}$, with Qwen3.5-122B-A10B serving as the teacher for offline competence estimation and token-level OPD.

\begin{figure*}[t]
    \centering
    \includegraphics[width=\textwidth]{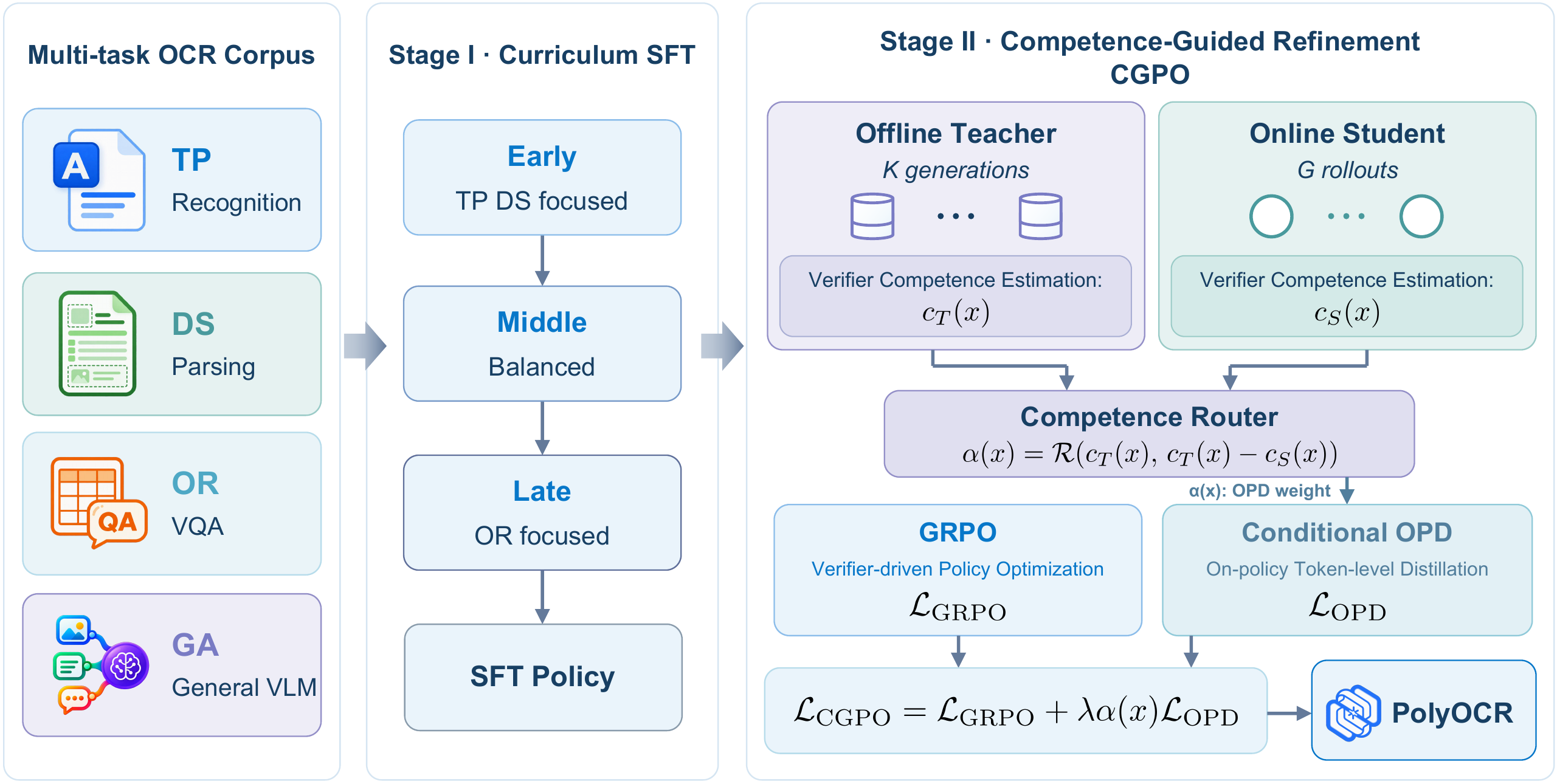}
    \caption{
    Overview of the \model{} training framework. Stage I applies a three-phase curriculum SFT schedule over Text Perception (TP), Document Structure Understanding (DS), OCR Understanding and Reasoning (OR), and General Multimodal Alignment (GA) data. Stage II performs CGPO by combining verifier-driven GRPO with conditional OPD. Here, $\mathcal{R}$ denotes the calibrated competence-routing map defined in Sec.~\ref{sec:cgpo}, which maps the cached teacher competence $c_T(x)$ and teacher--student competence gap $c_T(x)-c_S(x)$ to the OPD weight $\alpha(x)$.
    }
    \label{fig:training_pipeline}
\end{figure*}

\subsection{Supervised Fine-Tuning}
\label{sec:sft}

The SFT corpus contains task families that differ substantially in data volume, output structure, and learning difficulty. Sampling them in their raw proportions would overrepresent the largest sources and provide limited control over the contribution of each capability. We therefore partition the corpus into four groups and assign their sampling quotas explicitly. Text Perception (TP) covers text recognition, grounding, position-aware VQA, and text spotting. Document Structure Understanding (DS) includes key-information extraction and mapping, handwritten answer extraction, document, chart, and table parsing, and formula recognition. OCR Understanding and Reasoning (OR) comprises classification, translation, counting, mathematical question answering, and other OCR-grounded reasoning tasks. General Multimodal Alignment (GA) retains broad image understanding and instruction-following data.

Across the approximately 60 million SFT instances, TP, DS, OR, and GA account for $32.5\%$, $30.0\%$, $22.5\%$, and $15.0\%$, respectively. This allocation gives most of the optimization budget to text perception and document structure, while retaining dedicated coverage for OCR-grounded semantics, reasoning, and general multimodal behavior.

\begin{table}[H]
    \centering
    \small
    \caption{Three-phase sampling schedule for SFT. Percentages denote the within-phase proportions of the four data groups.}
    \label{tab:sft_sampling_schedule}
    \renewcommand{\arraystretch}{1.08}
    \setlength{\tabcolsep}{5pt}
    \begin{tabular}{@{}lccccc@{}}
        \toprule
        Phase & Instances & TP & DS & OR & GA \\
        \midrule
        Early  & 20M & $40.0\%$ & $35.0\%$ & $10.0\%$ & $15.0\%$ \\
        Middle & 20M & $32.5\%$ & $30.0\%$ & $22.5\%$ & $15.0\%$ \\
        Late   & 20M & $25.0\%$ & $25.0\%$ & $35.0\%$ & $15.0\%$ \\
        \midrule
        Overall & 60M & $32.5\%$ & $30.0\%$ & $22.5\%$ & $15.0\%$ \\
        \bottomrule
    \end{tabular}
\end{table}

We apply a three-phase quota schedule, with each phase containing 20 million instances. The early phase places greater weight on TP and DS. The OR share then increases from $10.0\%$ to $35.0\%$ as training progresses, whereas GA remains fixed at $15.0\%$. Each group is sampled without replacement according to its phase-specific quota, and the selected instances are shuffled within each phase. Consequently, the complete schedule traverses the full SFT corpus while preserving the overall mixture shown in Table~\ref{tab:sft_sampling_schedule}. The resulting checkpoint initializes the subsequent CGPO stage.

\subsection{Competence-Guided Policy Optimization}
\label{sec:cgpo}

The SFT stage establishes broad OCR competence, but its imitation objective does not directly optimize verifier-defined task outcomes. We therefore use outcome-supervised Group Relative Policy Optimization (GRPO), which assigns each response a scalar reward and derives token advantages from its relative reward within a rollout group~\citep{shao2024deepseekmath}. Our ablation in Table~\ref{tab:posttraining_ablation} reports a CC-OCR score of $81.72$ for GRPO and $82.18$ after adding fixed-weight OPD, supporting the use of complementary supervision.

Consider an illustrative long-text OCR case: a financial document contains 1,000 characters after normalization, including the amount \texttt{18,650}. Suppose the eight student rollouts preserve the surrounding text but transcribe this amount as \texttt{18,850} or \texttt{18,350}. Each output contains one character substitution, so normalized edit similarity assigns all eight responses the same reward, $1-1/1000=0.999$. The amount is incorrect despite the high page-level score. With $A_i=(r_i-\bar r)/(\operatorname{std}(r)+\epsilon)$, all group-relative advantages are zero, and this group contributes no reward-driven policy gradient to correct the digit. This equal-reward degeneracy is also discussed in DAPO~\citep{yu2025dapo}. Even when rewards differ, the response-level advantage does not identify the erroneous position.

On-policy distillation (OPD) supplies dense token-level supervision by evaluating the teacher on student-generated prefixes~\citep{agarwal2024opd}. In the example, a teacher distribution that favors the correct continuation when the amount is generated provides information that the tied outcome rewards cannot supply. However, this supervision inherits the teacher's errors: if the teacher favors \texttt{18,850}, distillation can reinforce the same incorrect digit. If the student later recognizes \texttt{18,650} correctly while the teacher still favors the wrong continuation, a fixed OPD weight can oppose the correction encouraged by the verifier. Teacher fluency and a high page-level score therefore do not establish the usefulness of its local guidance.

We propose \textit{Competence-Guided Policy Optimization} (CGPO) to retain GRPO as the optimization backbone and condition the OPD weight on both teacher quality and the teacher's advantage over the current student. Reliable teacher advantages warrant stronger distillation; unreliable guidance or a diminishing advantage warrants less. This sample-wise allocation preserves verifier-guided exploration as student competence changes.

For each training sample $x$ from task family $\tau$, let $y^\star$ denote its quality-verified reference. A task-specific verifier assigns an outcome reward $R_{\tau}(x,y,y^\star)\in[0,1]$. The verifier is instantiated according to the semantic structure of each task, using normalized edit similarity for text recognition, intersection-over-union (IoU) for visual grounding, character detection matching (CDM) for formula recognition~\citep{wang2024cdm}, tree-edit-distance-based similarity (TEDS) for table parsing~\citep{zhong2020pubtabnet}, field-level F1 for key information extraction, and an LLM-based correctness judge for open-ended OCR-centric VQA. Task-specific rewards also underpin multi-task OCR reinforcement learning in Infinity-Parser2~\citep{huang2026infinityparser2}, while criterion-based language-model assessment has been studied in G-Eval~\citep{liu2023geval}. Our task-specific metrics are mapped to a common reward space and supplied to GRPO for outcome-based policy optimization. The same verified reference and verifier are used to score teacher and student generations.

\paragraph{Competence estimation.}
Before CGPO optimization, the teacher is evaluated on each training sample by sampling $K$ responses:
\begin{equation}
    r_{T,k}(x)
    =
    R_{\tau}\!\left(x,y_T^{(k)},y^\star\right),
    \qquad
    c_T(x)
    =
    \frac{1}{K}
    \sum_{k=1}^{K}
    r_{T,k}(x),
\end{equation}
where $c_T(x)$ provides a verifier-based estimate of the teacher's competence on sample $x$. This quantity is computed before policy optimization and cached throughout training.

During CGPO optimization, the current student policy generates $G$ responses for the same sample, yielding an online estimate of student competence:
\begin{equation}
    r_{S,i}(x)
    =
    R_{\tau}\!\left(x,y_S^{(i)},y^\star\right),
    \qquad
    c_S(x)
    =
    \frac{1}{G}
    \sum_{i=1}^{G}
    r_{S,i}(x).
\end{equation}
The teacher--student competence gap is then defined as:
\begin{equation}
    \Delta(x)=c_T(x)-c_S(x).
\end{equation}
A positive $\Delta(x)$ indicates that the teacher remains ahead of the student on the current sample, whereas a small or negative value indicates that the student has approached or exceeded the teacher's competence. Importantly, a positive mean gap alone is insufficient to establish useful supervision: a low-quality teacher may still outperform a weaker student, and a gap estimated from finitely many rollouts may not reproduce under an independent draw.

\paragraph{Offline routing calibration.}
We determine the routing map on a fixed calibration split drawn from the quality-verified training pool using task- and difficulty-stratified sampling. For each calibration sample, teacher and SFT-student generations are divided into independent sets $A_T,B_T$ and $A_S,B_S$. The $A$ sets emulate the finite-rollout statistics available to the router,
\begin{equation}
    \widehat c_T^A(x)
    =\frac{1}{|A_T|}\sum_{k\in A_T}r_{T,k}(x),
    \qquad
    \widehat c_S^A(x)
    =\frac{1}{|A_S|}\sum_{i\in A_S}r_{S,i}(x),
    \qquad
    \widehat\Delta^A(x)=\widehat c_T^A(x)-\widehat c_S^A(x),
\end{equation}
where the sizes of the $A$ sets match the teacher and student rollout budgets used during CGPO. The independent $B$ sets provide two calibration targets. The first is held-out teacher quality,
\begin{equation}
    q_T^B(x)
    =\frac{1}{|B_T|}\sum_{k\in B_T}r_{T,k}(x),
\end{equation}
and the second is the tie-adjusted probability of superiority~\citep{vargha2000superiority}, estimated from all teacher--student pairs,
\begin{equation}
    p_{\mathrm{win}}^B(x)
    =
    \frac{1}{|B_T||B_S|}
    \sum_{k\in B_T}\sum_{i\in B_S}
    \left[
        \mathbb{I}\!\left(r_{T,k}>r_{S,i}\right)
        +\frac{1}{2}\mathbb{I}\!\left(r_{T,k}=r_{S,i}\right)
    \right],
\end{equation}
where a tie contributes one half. Thus, $p_{\mathrm{win}}^B=0.5$ indicates balanced empirical teacher wins and losses, whereas values above $0.5$ indicate an empirical tendency for the teacher to outperform the student.

We use the affine-sigmoid form commonly employed in held-out post-hoc calibration~\citep{guo2017calibration}. The maps take verifier-based competence estimates as inputs and predict independently estimated teacher quality and superiority. We fit two monotone one-dimensional maps,
\begin{equation}
    \widehat q_T^A(x)
    =\sigma\!\left(a_T\widehat c_T^A(x)+b_T\right),
    \qquad
    \widehat p_{\mathrm{win}}^A(x)
    =\sigma\!\left(a_{\Delta}\widehat\Delta^A(x)+b_{\Delta}\right),
\end{equation}
by minimizing task-balanced soft-label binary cross-entropy,
\begin{equation}
    \mathcal{L}_{\mathrm{cal}}
    =
    \sum_{x\in\mathcal{D}_{\mathrm{cal}}}w_x
    \left[
        \ell_{\mathrm{BCE}}\!\left(q_T^B(x),\widehat q_T^A(x)\right)
        +
        \ell_{\mathrm{BCE}}\!\left(p_{\mathrm{win}}^B(x),\widehat p_{\mathrm{win}}^A(x)\right)
    \right],
\end{equation}
where $\ell_{\mathrm{BCE}}(u,v)=-u\log v-(1-u)\log(1-v)$ and $w_x\propto 1/N_{\tau(x)}$ equalizes the aggregate contribution of task families, with $N_{\tau(x)}$ denoting the number of calibration samples from the task family of $x$. The slopes are constrained to be positive, and repeated independent partitions reduce sensitivity to a particular split. This calibration requires only offline inference and does not update either the teacher or the student. Once estimated, $a_T,b_T,a_{\Delta},b_{\Delta}$ are fixed throughout CGPO.

\paragraph{Calibrated competence routing.}
During CGPO, the frozen maps are evaluated using the cached teacher competence and current online student competence:
\begin{equation}
    \widehat q_T(x)
    =\sigma\!\left(a_Tc_T(x)+b_T\right),
    \qquad
    \widehat p_{\mathrm{win}}(x)
    =\sigma\!\left(a_{\Delta}\Delta(x)+b_{\Delta}\right).
\end{equation}
Here, $\widehat q_T(x)$ estimates the teacher's independent outcome quality, whereas $\widehat p_{\mathrm{win}}(x)$ estimates how reproducibly the teacher outperforms the current student. We define the sample-wise routing coefficient as
\begin{equation}
    \boxed{
    \alpha(x)
    =
    \operatorname{sg}\!\left[
        \widehat q_T(x)
        \left(2\widehat p_{\mathrm{win}}(x)-1\right)_{+}
    \right]},
\end{equation}
where $(z)_{+}=\max(z,0)$ and $\operatorname{sg}$ stops gradients through the routing decision. The excess-win term is zero whenever the teacher is no more likely than the student to succeed and increases from zero to one as the calibrated win probability rises from $0.5$ to $1$. Multiplication by $\widehat q_T(x)$ further suppresses teachers that beat the student only because both policies perform poorly. Consequently, $\alpha(x)\in[0,1]$ acts as a calibrated soft conjunction of absolute teacher quality and reproducible marginal advantage.

Routing operates at the sample level; local teacher errors can remain in samples with high aggregate rewards. The contribution and observed behavior of competence routing are examined in Sec.~\ref{sec:training_ablation} and Fig.~\ref{fig:cgpo_routing_behavior}.

\paragraph{On-policy token-level supervision.}
For a student-generated response $y_S$, OPD evaluates the teacher along the trajectory visited by the current policy. At decoding step $t$, the corresponding student prefix is $h_t=(x,q,y_{S,<t})$, and the token-level distillation objective is defined as
\begin{equation}
    \mathcal{L}_{\mathrm{OPD}}(x,y_S)
    =
    \frac{1}{|y_S|}
    \sum_{t=1}^{|y_S|}
    D_{\mathrm{KL}}\!\left(
    p_{\theta}(\cdot\mid h_t)
    \,\Vert\,
    p_T(\cdot\mid h_t)
    \right),
\end{equation}
where $D_{\mathrm{KL}}(p_{\theta}\Vert p_T)$ is the reverse KL divergence from the student token distribution to the teacher token distribution, a direction studied for generative-model distillation in MiniLLM and GKD~\citep{gu2024minillm,agarwal2024opd}. The teacher remains frozen during OPD, and gradients are taken only with respect to $\theta$. Since the teacher is queried on student-generated prefixes, the resulting supervision is aligned with the states visited by the evolving policy rather than with an independently generated teacher trajectory.

\paragraph{Joint optimization.}
The final CGPO objective is
\begin{equation}
    \mathcal{L}_{\mathrm{CGPO}}(x)
    =
    \mathcal{L}_{\mathrm{GRPO}}(x)
    +
    \lambda\,\alpha(x)\,
    \mathcal{L}_{\mathrm{OPD}}(x),
\end{equation}
where $\lambda$ controls the overall strength of teacher supervision. We do not include a reference-policy KL penalty in $\mathcal L_{\mathrm{GRPO}}$; the reverse KL in $\mathcal L_{\mathrm{OPD}}$ is used for teacher distillation. GRPO remains active throughout the CGPO stage and provides outcome-level optimization, while OPD contributes dense token-level supervision according to the competence-aware routing coefficient.

The resulting optimization exhibits an adaptive transition of supervision across training. For samples where the teacher initially has high calibrated quality and a reproducible advantage, CGPO assigns greater weight to OPD and exploits the teacher's fine-grained token-level knowledge. As the student improves, $c_S(x)$ increases, $\Delta(x)$ decreases, and the calibrated win probability approaches or falls below $0.5$, automatically suppressing distillation. Conversely, samples that remain substantially beyond the student's competence continue to receive teacher guidance in proportion to the teacher's independently calibrated quality. CGPO therefore follows the evolving teacher--student competence boundary throughout optimization, replacing a static distillation schedule with sample-adaptive supervision.

\section{Evaluation}
\label{sec:evaluation}

We conduct a comprehensive empirical evaluation of \model across diverse text-centric visual intelligence scenarios. The evaluation examines the model's ability to generalize beyond isolated OCR tasks, covering text perception, spatially grounded understanding, structured document analysis, information extraction, and OCR-oriented reasoning. We further investigate the effectiveness of the proposed Competence-Guided Policy Optimization (CGPO) framework.

\subsection{Evaluation Protocol}
\label{sec:eval_protocol}

We establish a unified evaluation protocol to assess \model across diverse OCR and document intelligence scenarios. All compared models are evaluated under consistent benchmark versions, inference configurations, and scoring procedures to ensure reliable and reproducible comparisons.

\paragraph{Models.}

We evaluate two model variants, \modelsmall and \modelbig. Both variants adopt the Qwen3.5 vision-language backbone and are trained under the unified OCR training framework described in Sections~\ref{sec:data} and~\ref{sec:training}. Unless otherwise specified, all evaluations are conducted under the same inference configuration, including prompt templates, image preprocessing strategies, visual token budgets, and decoding settings. Additionally, we include \model{}-2.7B in the MDPBench evaluation to assess multilingual document parsing capabilities. This variant features a specialized architecture comprising a 1.2B-parameter ViT vision encoder and a 1.5B-parameter Qwen2.5 text decoder, optimized for diverse writing systems and acquisition conditions.

\paragraph{OCRBench v2.1 protocol.}
We refine the original OCRBench v2 evaluation protocol to improve the reliability, consistency, and discriminative capability of benchmark evaluation. Due to the limited quality of some original reference annotations and the insufficient alignment between certain evaluation metrics and the actual task objectives, we revise two key aspects of the protocol: reference annotation quality and task-specific evaluation criteria. Specifically, reference annotations are systematically audited and corrected to reduce annotation noise, while task-specific evaluation procedures are refined to better align scoring mechanisms with the semantic objectives of individual tasks.

The resulting OCRBench v2.1 protocol preserves the original task coverage while providing more reliable annotations and more task-aligned evaluation criteria. Detailed modifications to annotations, metrics, and evaluation procedures are presented in Appendix~\ref{app:ocrbenchv21}. All OCRBench v2.1 results reported in this work are obtained using the refined evaluation protocol.

\paragraph{Benchmarks and baselines.} The evaluation suite consists of five representative benchmarks: OCRBench v2.1 (our revision of OCRBench v2~\citep{ocrbenchv2}), CC-OCR~\citep{yang2025ccocr}, OmniDocBench v1.6~\citep{ouyang2025omnidocbench}, MDPBench~\citep{li2026mdpbenchbenchmarkmultilingualdocument}, and an in-house KIE benchmark. OCRBench v2.1 and CC-OCR provide comprehensive evaluations across diverse OCR scenarios, covering broad text-centric visual capabilities. OmniDocBench v1.6 evaluates document-centric parsing ability, including layout analysis, table understanding, formula recognition, and document-level content reconstruction. MDPBench further extends the evaluation to multilingual document parsing, covering diverse language/script groups and both digital and photographed documents. The in-house KIE benchmark evaluates the model's ability to extract structured information from document images, focusing on field-level information extraction under practical application scenarios.

We compare \model{} against a diverse set of representative vision-language models, including both general-purpose and OCR-specialized models with different scales and architectural designs. For public benchmarks, we use their official test sets and retain the corresponding evaluation protocols for CC-OCR, OmniDocBench v1.6, and MDPBench. For OCRBench v2.1, all comparisons are conducted using the refined annotations and scoring procedures introduced above; therefore, scores obtained under the original OCRBench v2 protocol are not directly comparable. For the in-house KIE benchmark, we evaluate models using the same test set and field-level F1 metric to ensure consistent comparison across different systems.

\subsection{Overall Comparison}
\label{sec:overall_results}

Table~\ref{tab:overall_results} summarizes the cross-benchmark performance of \model against representative general-purpose vision-language models and OCR-specialized systems. The evaluated benchmarks collectively assess a broad spectrum of text-centric visual capabilities, including fundamental text perception, document-level structural understanding, and OCR-oriented semantic reasoning. \modelbig demonstrates competitive and consistent performance across all evaluated benchmarks. It achieves scores of \textbf{80.42} and \textbf{78.37} on the English and Chinese settings of OCRBench v2.1, respectively, and obtains a score of \textbf{82.43} on CC-OCR. Furthermore, \modelbig reaches \textbf{91.57} on OmniDocBench v1.6, demonstrating its capability to extend from fine-grained text recognition toward complex document structure understanding and parsing.

The cross-benchmark evaluation demonstrates that the proposed model maintains robust performance across heterogeneous OCR scenarios and task formulations. Such capability consistency reflects the effectiveness of the unified training paradigm in developing a general-purpose OCR foundation model with broad text-centric visual understanding abilities. Its competitive performance across diverse task settings, including text transcription, spatial localization, structured document parsing, and semantic reasoning, highlights the model's ability to address varied OCR requirements while maintaining balanced capabilities across different levels of text-centric visual intelligence. The following sections further analyze the model performance on individual benchmarks and provide detailed comparisons across different OCR capabilities.

 \begin{table*}[t]
      \centering
      \small
      \setlength{\tabcolsep}{10pt}
      \caption{
      Cross-benchmark comparison of \model{} with representative vision-language models. 
OCRBench v2.1 reports English/Chinese scores. 
Bold and underline indicate the best and second-best results, respectively.
``--'' indicates models not evaluated on the corresponding benchmark in this report.
      }
      \label{tab:overall_results}
      \begin{tabular*}{\textwidth}{@{}l@{\extracolsep{\fill}\hspace{5em}}ccc@{}}
          \toprule
          Model &
          OCRBench v2.1 (en/zh) $\uparrow$ &
          CC-OCR $\uparrow$ &
          OmniDocBench v1.6 $\uparrow$ \\
          \midrule

          \multicolumn{4}{@{}l}{\textit{General-purpose Vision-Language Models}} \\
          \hspace{1.5em}\modelicon{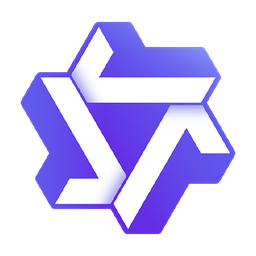}Qwen3.5-2B           & 54.03/63.75 & 65.66 & 79.95 \\
          \hspace{1.5em}\modelicon{assets/model_logos/table/qwen.png}Qwen3.5-9B           & 66.64/71.88 & 71.44 & 89.49 \\
          \hspace{1.5em}\modelicon{assets/model_logos/table/qwen.png}Qwen3-VL-2B-Instruct & 44.27/48.84 & 51.02 & 80.05 \\
          \hspace{1.5em}\modelicon{assets/model_logos/table/qwen.png}Qwen3-VL-8B-Instruct & 54.36/70.56 & 70.29 & 78.94 \\
          \hspace{1.5em}\modelicon{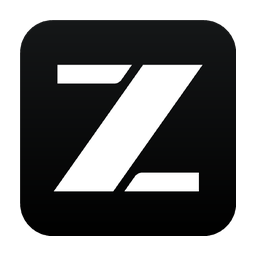}GLM-4.6V-Flash       & 67.89/75.34 & 72.50 & 71.42 \\
          \hspace{1.5em}\modelicon{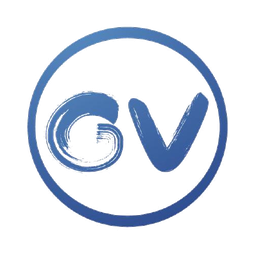}InternVL3.5-8B       & 51.09/62.88 & 65.88 & 75.89 \\
          \hspace{1.5em}\modelicon{assets/model_logos/table/qwen.png}Qwen3.6-35B-A3B      & 70.34/76.25 & 73.19 & 89.01 \\
          \hspace{1.5em}\modelicon{assets/model_logos/table/qwen.png}Qwen3.5-122B-A10B    & 70.34/\underline{78.77} & 73.79 & 89.04 \\
          \hspace{1.5em}\modelicon{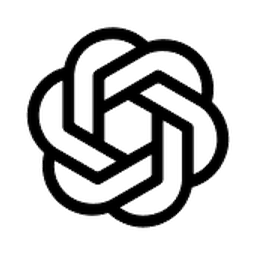}GPT-5.6-Sol          & 72.92/67.80 & 74.54 & 88.00 \\
          \hspace{1.5em}\modelicon{assets/model_logos/table/qwen.png}Qwen3.8-Max          & \underline{77.75}/\textbf{79.32} & 73.64 & \underline{92.31} \\

          \midrule
          \multicolumn{4}{@{}l}{\textit{OCR-specialized Vision-Language Models}} \\
          \hspace{1.5em}\modelicon{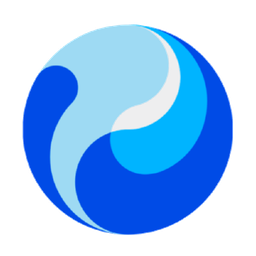}HunyuanOCR      & 32.26/54.44 & 54.57 & 87.92 \\
          \hspace{1.5em}\modelicon{assets/model_logos/table/hunyuan.png}HunyuanOCR-1.5  & 32.63/49.41 & 51.10 & 92.17 \\
          \hspace{1.5em}\modelicon{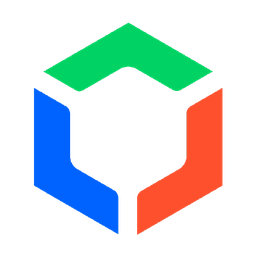}Qianfan-OCR     & 57.09/70.87 & \underline{82.06} & 88.37 \\
          \hspace{1.5em}\modelicon{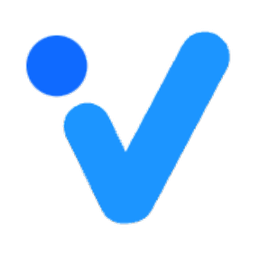}OvisOCRv2       & 25.67/50.28 & 36.26 & \textbf{96.58} \\

          \midrule
          \multicolumn{4}{@{}l}{\textit{\model{} (Ours)}} \\
          \hspace{1.5em}\ourmodelicon\textbf{\model{}-2B} & 72.33/71.32 & 80.53 & 88.12 \\
          \hspace{1.5em}\ourmodelicon\textbf{\model{}-9B} & \textbf{80.42}/78.37 & \textbf{82.43} & 91.57 \\

          \bottomrule
      \end{tabular*}
  \end{table*}

\subsection{OCRBench v2.1 Evaluation}
\label{sec:ocrbenchv2_results}

OCRBench v2.1 provides a comprehensive evaluation of text-centric visual intelligence across diverse OCR capabilities, including text recognition, localization, spotting, document parsing, formula understanding, translation, and OCR-oriented reasoning. Table~\ref{tab:ocrbench_subcategories} presents a detailed comparison across these capability categories, providing a fine-grained assessment of model performance beyond overall accuracy.

\modelbig achieves competitive performance on OCRBench v2.1, obtaining overall scores of \textbf{80.42} and \textbf{78.37} on the English and Chinese settings, respectively. The model demonstrates strong results across text perception, document understanding, and OCR-oriented reasoning. These results indicate that the proposed unified training framework effectively enables the model to handle heterogeneous OCR tasks within a single generative modeling paradigm.

At the category level, \modelbig ranks first among the compared models in five of the 13 language-specific categories: English text recognition ($82.99$), text detection ($70.84$), element parsing ($89.70$), and visual text understanding ($87.58$), together with Chinese element parsing ($86.86$). It ranks second in English text spotting ($78.50$) and English relationship extraction ($95.51$). These results show strengths in text perception and structured parsing, while mathematical calculation, Chinese text recognition, and Chinese visual text understanding remain below the strongest evaluated baselines.

Overall, the OCRBench v2.1 results demonstrate that \model provides a balanced and comprehensive OCR capability across heterogeneous task formulations, establishing a strong foundation for unified text-centric visual intelligence.

\begin{table*}[t]
  \centering
  \footnotesize 
  \setlength{\tabcolsep}{2pt} 
  \caption{
Fine-grained OCRBench v2.1 comparison of \model{} with representative vision-language models.  
TR, TD, TS, RE, EP, MC, VTU, and KR denote Text Recognition, Text Detection, Text Spotting, Relationship Extraction, Element Parsing, Mathematical Calculation, Visual Text Understanding, and Knowledge Reasoning, respectively. 
The \textbf{All} column reports overall English/Chinese scores (en/zh). 
Bold and underline indicate the best and second-best results.
}
  \label{tab:ocrbench_subcategories}
  \begin{tabular*}{\textwidth}{@{}l@{\hspace{4pt}\extracolsep{\fill}}cccccccccccccc@{}}
      \toprule
      & \multicolumn{8}{c}{English} & \multicolumn{5}{c}{Chinese} & \\
      \cmidrule(lr){2-9} \cmidrule(lr){10-14}
      Model & TR & TD & TS & RE & EP & MC & VTU & KR & TR & RE & EP & VTU & KR & All (en/zh) \\
      \midrule

      \multicolumn{15}{@{}l}{\textit{General-purpose VLM}} \\
      \hspace{1em}\modelicon{assets/model_logos/table/qwen.png}Qwen3.5-2B           & 67.44 & 31.48 & 8.50  & 80.13 & 58.73 & 52.65 & 71.73 & 61.58 & 73.48 & 71.67 & 66.15 & 60.50 & 46.95 & 54.03/63.75 \\
      \hspace{1em}\modelicon{assets/model_logos/table/qwen.png}Qwen3.5-9B           & 72.16 & 54.42 & 49.94 & 89.44 & 66.12 & 60.45 & 76.23 & 64.37 & 76.42 & 78.62 & 67.35 & 65.50 & 71.52 & 66.64/71.88 \\
      \hspace{1em}\modelicon{assets/model_logos/table/qwen.png}Qwen3-VL-2B-Instruct & 51.83 & 24.09 & 0.64  & 73.49 & 49.58 & 29.89 & 70.44 & 54.19 & 46.55 & 67.90 & 50.81 & 50.00 & 28.93 & 44.27/48.84 \\
      \hspace{1em}\modelicon{assets/model_logos/table/qwen.png}Qwen3-VL-8B-Instruct & 67.71 & 26.94 & 4.20  & 88.54 & 64.76 & 45.11 & 74.85 & 62.79 & 77.75 & 83.91 & 72.26 & \underline{68.00} & 50.87 & 54.36/70.56 \\
      \hspace{1em}\modelicon{assets/model_logos/table/glm.png}GLM-4.6V-Flash       & 78.04 & 54.08 & 38.02 & 83.21 & 68.13 & 71.10 & 82.14 & 68.44 & 76.78 & 84.53 & 77.60 & 62.00 & 75.79 & 67.89/75.34 \\
      \hspace{1em}\modelicon{assets/model_logos/table/internvl.png}InternVL3.5-8B       & 60.96 & 27.45 & 5.89  & 83.79 & 54.12 & 46.25 & 70.95 & 59.33 & 63.00 & 70.38 & 69.17 & 53.00 & 58.84 & 51.09/62.88 \\
      \hspace{1em}\modelicon{assets/model_logos/table/qwen.png}Qwen3.6-35B-A3B      & 75.15 & 58.32 & 67.13 & 92.36 & 64.23 & 61.83 & 77.13 & 66.58 & 75.60 & 85.95 & 72.35 & 66.00 & 81.34 & 70.34/76.25 \\
      \hspace{1em}\modelicon{assets/model_logos/table/qwen.png}Qwen3.5-122B-A10B    & 78.63 & 58.53 & 62.24 & 95.22 & 68.96 & 59.17 & 76.49 & 63.50 & 80.39 & \underline{87.07} & 79.65 & 66.50 & 80.25 & 70.34/\underline{78.77} \\
      \hspace{1em}\modelicon{assets/model_logos/table/openai.png}GPT-5.6-Sol          & 74.21 & 61.48 & 67.45 & 88.55 & 65.87 & \underline{76.73} & 76.59 & \underline{72.46} & 70.10 & 71.90 & 74.06 & 51.00 & 71.95 & 72.92/67.80 \\
      \hspace{1em}\modelicon{assets/model_logos/table/qwen.png}Qwen3.8-Max          & \underline{81.52} & \underline{70.04} & \textbf{78.62} & \textbf{96.58} & 69.36 & 75.47 & 77.23 & \textbf{73.16} & \underline{80.73} & \textbf{89.86} & 73.98 & \underline{68.00} & \underline{84.04} & \underline{77.75}/\textbf{79.32} \\
      \hspace{1em}\modelicon{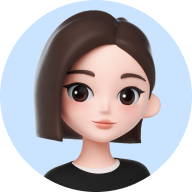}Doubao-Seed-2.0-Lite & 71.35 & 66.91 & 62.17 & 90.62 & 65.03 & \textbf{82.66} & \underline{83.43} & 72.12 & 59.89 & 83.34 & 71.43 & 62.50 & \textbf{93.40} & 74.28/74.11 \\
      \hspace{1em}\modelicon{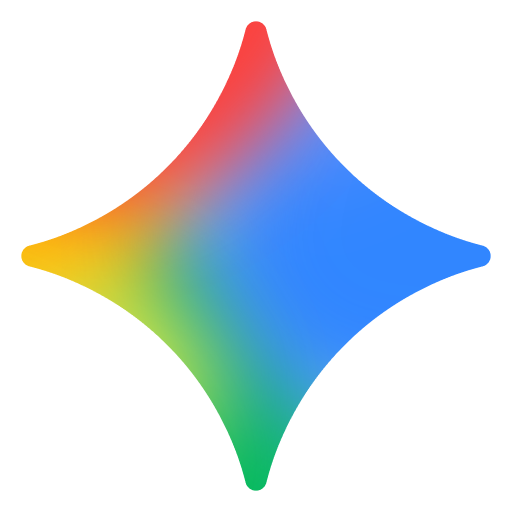}Gemini-3.5-Flash-Lite & 72.71 & 24.94 & 0.50 & 92.30 & 65.23 & 49.22 & 76.22 & 72.20 & 37.74 & 79.66 & 54.61 & 66.50 & 51.51 & 56.67/58.01 \\
      \hspace{1em}\modelicon{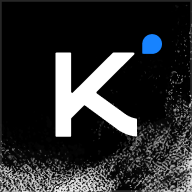}Kimi-K3 & 77.20 & 57.25 & 45.77 & 91.65 & 65.02 & 58.93 & 75.88 & 70.56 & 62.36 & 80.24 & 67.07 & 62.50 & 81.59 & 67.78/70.75 \\

      \midrule
      \multicolumn{15}{@{}l}{\textit{OCR-specialized VLM}} \\
      \hspace{1em}\modelicon{assets/model_logos/table/hunyuan.png}HunyuanOCR      & 72.26 & 5.50  & 0.00 & 19.84 & 44.00 & 12.25 & 63.11 & 41.13 & 75.86 & 62.59 & 60.18 & 48.81 & 24.76 & 32.26/54.44 \\
      \hspace{1em}\modelicon{assets/model_logos/table/hunyuan.png}HunyuanOCR-1.5  & 71.77 & 14.99 & 0.33 & 1.11  & 48.16 & 14.55 & 65.39 & 44.70 & 75.73 & 21.98 & 61.26 & 55.29 & 32.79 & 32.63/49.41 \\
      \hspace{1em}\modelicon{assets/model_logos/table/qianfan.png}Qianfan-OCR     & 73.56 & 40.04 & 1.86 & 90.82 & 77.42 & 31.84 & 81.53 & 59.70 & \textbf{80.78} & 85.24 & 86.09 & 62.50 & 39.74 & 57.09/70.87 \\
      \hspace{1em}\modelicon{assets/model_logos/table/ovis.png}OvisOCRv2       & 57.19 & 0.00  & 0.00 & 0.85  & 43.31 & 17.90 & 53.93 & 32.14 & 73.00 & 29.72 & 65.89 & \textbf{69.00} & 13.81 & 25.67/50.28 \\

      \midrule
      \multicolumn{15}{@{}l}{\textit{\model{} (Ours)}} \\
      \hspace{1em}\ourmodelicon\model{}-2B & 77.90 & 60.90 & 60.11 & 94.06 & \underline{89.30} & 53.09 & 81.47 & 61.79 & 77.26 & 79.39 & \underline{86.34} & 52.50 & 61.09 & 72.33/71.32 \\
      \hspace{1em}\ourmodelicon\model{}-9B & \textbf{82.99} & \textbf{70.84} & \underline{78.50} & \underline{95.51} & \textbf{89.70} & 68.15 & \textbf{87.58} & 70.11 & 77.13 & 85.38 & \textbf{86.86} & 60.50 & 82.00 & \textbf{80.42}/78.37 \\

      \bottomrule
  \end{tabular*}
\end{table*}

\subsection{CC-OCR Evaluation}
\label{sec:ccocr_results}

CC-OCR provides an evaluation of practical OCR capabilities under diverse visual conditions, covering real-world text reading, document-level understanding, structured information extraction, and multilingual OCR. Our CC-OCR evaluation covers Multi-Scene OCR, Document Parsing, and Key Information Extraction (KIE); the multilingual track is excluded.

Table~\ref{tab:ccocr} presents the comparison across the three evaluated tracks. \modelbig achieves an overall score of \textbf{82.43}, demonstrating competitive performance across practical OCR scenarios. In particular, the model obtains the best results on Document
Parsing and Key Information Extraction, reaching scores of \textbf{71.71} and \textbf{94.02}, respectively. These results indicate that the proposed unified training framework effectively supports both fine-grained text understanding and document-level structured prediction.

The strong performance on KIE further highlights the advantage of jointly modeling text perception and semantic information extraction within a unified OCR framework. \model learns to associate visual text regions with semantic structures and output formats, supporting structured information extraction from complex documents. Together with the Document Parsing results, the CC-OCR evaluation demonstrates the capability of \model to extend beyond text recognition toward structured text-centric visual intelligence.

\begin{table*}[t]
    \centering
      \small
      \setlength{\tabcolsep}{12pt}
      \caption{
      CC-OCR results of \model{} and representative general-purpose and
      OCR-specialized vision-language models. The overall score is the average of Document Parsing, Key Information Extraction, and Multi-Scene OCR. Bold and underline indicate the best and second-best results, respectively.
      }
      \label{tab:ccocr}
      \begin{tabular*}{\textwidth}{@{}l@{\extracolsep{\fill}}cccc@{}}
          \toprule
          Model & Doc Parsing $\uparrow$ & KIE $\uparrow$ & Multi-Scene OCR $\uparrow$ & Overall $\uparrow$ \\
          \midrule

          \multicolumn{5}{@{}l}{\textit{General-purpose Vision-Language Models}} \\
          \hspace{1.5em}\modelicon{assets/model_logos/table/qwen.png}Qwen3.5-2B           & 32.48 & 83.01 & 81.49 & 65.66 \\
          \hspace{1.5em}\modelicon{assets/model_logos/table/qwen.png}Qwen3.5-9B           & 37.95 & 91.40 & 84.98 & 71.44 \\
          \hspace{1.5em}\modelicon{assets/model_logos/table/qwen.png}Qwen3-VL-2B-Instruct & 16.38 & 81.48 & 55.21 & 51.02 \\
          \hspace{1.5em}\modelicon{assets/model_logos/table/qwen.png}Qwen3-VL-8B-Instruct & 38.89 & 91.81 & 80.16 & 70.29 \\
          \hspace{1.5em}\modelicon{assets/model_logos/table/glm.png}GLM-4.6V-Flash       & 49.43 & 83.12 & 84.94 & 72.50 \\
          \hspace{1.5em}\modelicon{assets/model_logos/table/internvl.png}InternVL3.5-8B       & 35.92 & 86.96 & 74.75 & 65.88 \\
          \hspace{1.5em}\modelicon{assets/model_logos/table/qwen.png}Qwen3.6-35B-A3B      & 41.46 & 92.31 & 85.80 & 73.19 \\
          \hspace{1.5em}\modelicon{assets/model_logos/table/qwen.png}Qwen3.5-122B-A10B    & 41.71 & \underline{93.72} & \underline{85.94} & 73.79 \\
          \hspace{1.5em}\modelicon{assets/model_logos/table/openai.png}GPT-5.6-Sol          & 51.42 & 89.33 & 82.87 & 74.54 \\
          \hspace{1.5em}\modelicon{assets/model_logos/table/qwen.png}Qwen3.8-Max          & 42.68 & 91.12 & \textbf{87.11} & 73.64 \\

          \midrule
          \multicolumn{5}{@{}l}{\textit{OCR-specialized Vision-Language Models}} \\
          \hspace{1.5em}\modelicon{assets/model_logos/table/hunyuan.png}HunyuanOCR      & 40.95 & 45.34 & 77.41 & 54.57 \\
          \hspace{1.5em}\modelicon{assets/model_logos/table/hunyuan.png}HunyuanOCR-1.5  & 68.63 & 0.55  & 84.13 & 51.10 \\
          \hspace{1.5em}\modelicon{assets/model_logos/table/qianfan.png}Qianfan-OCR     & 71.18 & 92.12 & 82.88 & \underline{82.06} \\
          \hspace{1.5em}\modelicon{assets/model_logos/table/ovis.png}OvisOCRv2       & 39.16 & 0.00  & 69.64 & 36.26 \\

          \midrule
          \multicolumn{5}{@{}l}{\textit{\model{} (Ours)}} \\
          \hspace{1.5em}\ourmodelicon\model{}-2B & \underline{71.50} & 90.90 & 79.18 & 80.53 \\
          \hspace{1.5em}\ourmodelicon\model{}-9B & \textbf{71.71} & \textbf{94.02} & 81.55 & \textbf{82.43} \\

          \bottomrule
      \end{tabular*}
  \end{table*}

\subsection{OmniDocBench v1.6 Evaluation}
\label{sec:omnidoc_results}

OmniDocBench v1.6 targets document parsing within complex page layouts, focusing on the evaluation of capabilities such as text recognition, formula recognition, and table parsing. Table~\ref{tab:omnidocbench} summarizes PolyOCR's performance across various document components and reading orders.
When processing full pages directly, \model achieved a composite score of \textbf{91.57}, demonstrating strong competitiveness compared to general-purpose vision-language models. This indicates that the recognition capabilities acquired through unified OCR training can be applied to page-level document parsing, enabling the model to extract diverse types of document content from complex layouts.

To further examine \model's parsing capabilities regarding various document components, we introduced a layout-guided pipeline setup. This configuration utilizes PP-DocLayoutV3 to detect layout elements and predict reading order; the same \model model then parses the cropped regions containing texts, formulas, and tables, and the results are finally assembled into a Markdown document based on the predicted reading order. By delegating page layout analysis to an external module, this setup allows \model to focus more effectively on content recognition and structural recovery within the specific regions.

Under this configuration, the composite score rose from \textbf{91.57} to \textbf{95.52}, with improvements observed across all reported metrics. Notably, the formula recognition CDM score increased from 91.71 to 96.30, and the table parsing TEDS score improved from 88.67 to 93.56. These results further validate \model's effectiveness as a unified document component parser capable of handling texts, formulas, and tables with a single model. When combined with a layout analysis module, \model achieves page-level parsing performance comparable to leading specialized systems.

\begin{table*}[t]
      \centering
      \small
      \setlength{\tabcolsep}{3pt}
      \caption{
      Fine-grained OmniDocBench v1.6 results of \model{} and representative
general-purpose and OCR-specialized vision-language models. Bold and underline indicate
the best and second-best results, respectively.
      }
      \label{tab:omnidocbench}
      \begin{tabular*}{\textwidth}{@{}l@{\extracolsep{\fill}}cccccc@{}}
          \toprule
          & \shortstack{Text\textsuperscript{Edit} $\downarrow$}
          & \shortstack{Formula\textsuperscript{CDM} $\uparrow$}
          & \shortstack{Table\textsuperscript{TEDS} $\uparrow$}
          & \shortstack{Table\textsuperscript{TEDS-S} $\uparrow$}
          & \shortstack{Read Order\textsuperscript{Edit} $\downarrow$}
          & \shortstack{Overall $\uparrow$} \\
          \midrule

          \multicolumn{7}{@{}l}{\textit{General-purpose VLM}} \\
          \hspace{1.5em}\modelicon{assets/model_logos/table/qwen.png}Qwen3.5-2B           & 0.149 & 86.92 & 67.79 & 72.60 & 0.231 & 79.95 \\
          \hspace{1.5em}\modelicon{assets/model_logos/table/qwen.png}Qwen3.5-9B           & 0.080 & 92.44 & 84.01 & 87.42 & 0.178 & 89.49 \\
          \hspace{1.5em}\modelicon{assets/model_logos/table/qwen.png}Qwen3-VL-8B-Instruct & 0.110 & 79.68 & 68.20 & 73.69 & 0.222 & 78.94 \\
          \hspace{1.5em}\modelicon{assets/model_logos/table/glm.png}GLM-4.6V-Flash       & 0.238 & 77.96 & 60.14 & 64.07 & 0.280 & 71.42 \\
          \hspace{1.5em}\modelicon{assets/model_logos/table/internvl.png}InternVL3.5-8B       & 0.175 & 79.18 & 66.02 & 72.20 & 0.241 & 75.89 \\
          \hspace{1.5em}\modelicon{assets/model_logos/table/qwen.png}Qwen3.6-35B-A3B      & 0.077 & 91.35 & 83.41 & 86.18 & 0.176 & 89.01 \\
          \hspace{1.5em}\modelicon{assets/model_logos/table/qwen.png}Qwen3.5-122B-A10B    & 0.063 & 92.23 & 81.18 & 84.06 & 0.161 & 89.04 \\
          \hspace{1.5em}\modelicon{assets/model_logos/table/openai.png}GPT-5.6-Sol          & 0.109 & 89.68 & 85.18 & 88.87 & 0.165 & 88.00 \\
          \hspace{1.5em}\modelicon{assets/model_logos/table/qwen.png}Qwen3.8-Max          & 0.054 & 94.49 & 87.81 & 90.65 & 0.138 & 92.31
          \\

          \midrule
          \multicolumn{7}{@{}l}{\textit{OCR-specialized VLM}} \\
          \hspace{1.5em}\modelicon{assets/model_logos/table/hunyuan.png}HunyuanOCR      & 0.073 & 87.15 & 83.92 & 86.62 & 0.152 & 87.92 \\
          \hspace{1.5em}\modelicon{assets/model_logos/table/hunyuan.png}HunyuanOCR-1.5  & 0.049 & 89.95 & 91.50 & 93.05 & 0.131 & 92.17 \\
          \hspace{1.5em}\modelicon{assets/model_logos/table/qianfan.png}Qianfan-OCR     & 0.105 & 92.68 & 82.96 & 85.70 & 0.168 & 88.37 \\
          \hspace{1.5em}\modelicon{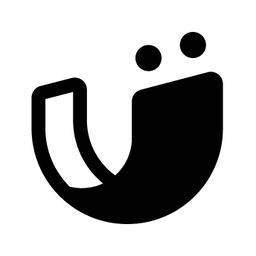}MinerU2.5-Pro   & 0.036 & 97.45 & 93.42 & 95.92 & \underline{0.120} & 95.75 \\
          \hspace{1.5em}\paddleicon PaddleOCR-VL-1.6 & 0.033 & \underline{97.49} & \underline{94.76} & 97.11 & 0.127 & 96.33 \\
          \hspace{1.5em}\modelicon{assets/model_logos/table/ovis.png}OvisOCRv2       & \textbf{0.025} & \textbf{97.53} & \underline{94.76} & \underline{97.16} & \textbf{0.111} & \underline{96.58} \\
          \hspace{1.5em}\modelicon{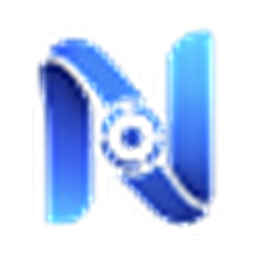}NaviDC-OCR      & \underline{0.027} & 96.36 & \textbf{97.05} & \textbf{98.52} & 0.122 & \textbf{96.87} \\

          \midrule
          \multicolumn{7}{@{}l}{\textit{\model{} (Ours)}} \\
          \hspace{1.5em}\ourmodelicon\model{}-9B & 0.057 & 91.71 & 88.67 & 91.80 & 0.139 & 91.57 \\
          \hspace{1.5em}\ourmodelicon\model{}-9B(Pipeline) & 0.033 & 96.30 & 93.56 & 95.38 & 0.123 & 95.52 \\

          \bottomrule
      \end{tabular*}
\end{table*}

\subsection{In-house KIE Benchmark Evaluation}
\label{sec:internal_kie}

We evaluate PolyOCR on an in-house KIE benchmark covering four image
types: Xuexin Network certificates, identity cards, admission notices,
and international passports. We report field-level F1 scores for each
image type as well as the overall test set performance.

As shown in Table~\ref{tab:polyocr_kie_combined}, PolyOCR-9B achieves the
highest overall F1 score of \textbf{93.60\%} among all evaluated models.
It achieves the best performance on admission notices, reaching an F1
score of \textbf{88.35\%}. On Xuexin Network certificates, PolyOCR-9B
obtains an F1 score of 95.40\%, slightly below PolyOCR-2B (95.53\%),
while both PolyOCR variants outperform all other evaluated models.
For identity cards, PolyOCR-9B achieves an F1 score of 99.20\%, which is
competitive with the best-performing model, Qwen3.6-35B-A3B (99.53\%).
On international passports, PolyOCR-9B reaches an F1 score of 93.80\%,
ranking second among the evaluated models and trailing GPT-5.6-Sol
(98.55\%) by 4.75 percentage points.

Overall, PolyOCR demonstrates strong and balanced performance across the
four image types. The PolyOCR variants achieve leading results on the
benchmark, with PolyOCR-9B showing particularly strong capabilities on
admission notices and maintaining competitive performance on other
complex KIE scenarios. Moreover, the overall F1 score of 93.60\%
achieved by PolyOCR-9B surpasses those of the evaluated larger-scale
models, including GPT-5.6-Sol (90.98\%), Qwen3.6-35B-A3B (89.17\%), and
Qwen3.5-122B-A10B (89.03\%). These results demonstrate the effectiveness
of PolyOCR in handling diverse KIE scenarios and highlight its potential
for practical document information extraction applications.

\begin{table*}[t]
\centering
\small
\setlength{\tabcolsep}{4pt}
\caption{Field-level F1 scores (\%) on the in-house KIE benchmark.
The benchmark contains four image types: Xuexin Network certificates,
identity cards, admission notices, and international passports.
The highest value in each column is shown in bold, and the second-highest
value is underlined.}
\label{tab:polyocr_kie_combined}
\begin{tabular*}{\textwidth}{@{}l@{\extracolsep{\fill}}ccccc@{}}
\toprule
\textbf{Model}
& \textbf{Xuexin}
& \textbf{Identity Card}
& \textbf{Admission Notice}
& \textbf{Passport}
& \textbf{Overall} \\
\midrule

\modelicon{assets/model_logos/table/openai.png}GPT-5.6-Sol
    & 93.03
    & 96.31
    & 79.58
    & \textbf{98.55}
    & 90.98 \\

\modelicon{assets/model_logos/table/qwen.png}Qwen3.6-35B-A3B
    & 89.19
    & \textbf{99.53}
    & 82.49
    & 89.54
    & 89.17 \\

\modelicon{assets/model_logos/table/qwen.png}Qwen3.5-122B-A10B
    & 92.33
    & \underline{99.43}
    & 79.27
    & 88.74
    & 89.03 \\

\modelicon{assets/model_logos/table/glm.png}GLM-4.6V-Flash
    & 90.83
    & 98.94
    & 74.35
    & 56.18
    & 79.71 \\

\modelicon{assets/model_logos/table/qianfan.png}Qianfan-OCR
    & 90.79
    & 99.38
    & 69.98
    & 59.78
    & 78.81 \\

\modelicon{assets/model_logos/table/internvl.png}InternVL3.5-8B
    & 87.47
    & 96.58
    & 61.57
    & 57.37
    & 74.66 \\

\modelicon{assets/model_logos/table/hunyuan.png}HunyuanOCR
    & 62.75
    & 99.38
    & 60.13
    & 31.34
    & 61.92 \\

\midrule
\ourmodelicon\model{}-2B
    & \textbf{95.53}
    & 99.17
    & \underline{88.17}
    & 93.41
    & \underline{93.49} \\

\ourmodelicon\model{}-9B
    & \underline{95.40}
    & 99.20
    & \textbf{88.35}
    & \underline{93.80}
    & \textbf{93.60} \\

\bottomrule
\end{tabular*}
\end{table*}

\subsection{MDPBench Evaluation}
\label{sec:multilingual_mdpbench}

MDPBench~\citep{li2026mdpbenchbenchmarkmultilingualdocument}
evaluates multilingual document parsing across digital and photographed
documents, covering 17 language/script groups with diverse writing
systems and acquisition conditions.
The benchmark uses $1-\mathrm{NED}$ for text recognition and reading
order, where NED denotes normalized edit distance, and CDM and TEDS
for formula and table recognition, respectively.
Scores are aggregated at the page level before averaging across pages.
Table~\ref{tab:realdoc_mdpbench_overall} reports overall performance,
acquisition-type breakdowns, and both group-level and individual
language/script scores.
All scores are expressed as percentages, with higher values indicating
better performance.

To assess multilingual document parsing within the \model{} family,
we additionally evaluate \model{}-2.7B, a specialized configuration
combining a 1.2B-parameter ViT vision encoder with a
1.5B-parameter Qwen2.5 text decoder.
The model is trained on an in-house document corpus curated through
cross-model consistency checks, multilingual quality filtering,
and validation of fields and reading order.

As shown in Table~\ref{tab:realdoc_mdpbench_overall},
\model{}-2.7B leads four of the five summary measures and ties for
the highest score on photographed documents among the listed methods.
It achieves an overall score of 87.0\%, exceeding GPT-5.6-Sol and
Gemini-3-Pro by 0.4 and 0.6 percentage points, respectively.
On digital documents, it reaches 91.0\%, surpassing the strongest
listed baseline, Kimi-K3, by 0.2 percentage points.
On photographed documents, it obtains 85.6\%, matching GPT-5.6-Sol.
Together, these results show competitive performance across both
acquisition types.

Compared with the listed OCR-specialized models, \model{}-2.7B
achieves higher scores across all five summary measures.
Relative to MonkeyOCRv2-B-Parsing, the OCR-specialized baseline
with the highest overall score, it improves overall performance
by 3.7 percentage points and photographed-document performance
by 3.9 percentage points.
The corresponding gains on the Latin and non-Latin averages
are 4.6 and 2.8 percentage points, respectively, indicating that
its advantage extends across both language/script groups.

The finer-grained results show Latin and non-Latin averages
of 88.8\% and 84.9\%, respectively, each exceeding the strongest
corresponding reference result by 0.4 percentage points.
At the individual language/script level, \model{}-2.7B achieves
the highest scores in six groups---Spanish, French, Dutch,
Vietnamese, Japanese, and Traditional Chinese---and ranks
among the top two in 12 of the 17 groups.
Overall, these results demonstrate strong multilingual document
parsing performance for the 2.7B configuration across diverse
writing systems and acquisition conditions.

\begin{table*}[t]
\centering
\small
\setlength{\tabcolsep}{2pt}

\caption{Fine-grained MDPBench scores (\%) of \model{}, general-purpose
VLMs, OCR-specialized VLMs, and document parsing pipelines.
All denotes the overall score; Digit.\ and Photo.\ report scores on
digital and photographed documents, respectively.
Latin and Non-Latin include group averages (Avg.) and individual
language/script scores. Higher scores indicate better performance.
Bold and underline indicate the highest and second-highest distinct
scores, respectively, in each column across all models and tools shown.
Tied scores receive the same formatting.}
\label{tab:realdoc_mdpbench_overall}

\resizebox{\textwidth}{!}{%
\begin{tabular}{@{}l*{22}{c}@{}}
\toprule
Model
& \multicolumn{3}{c}{Overall}
& \multicolumn{10}{c}{Latin}
& \multicolumn{9}{c}{Non-Latin} \\
\cmidrule(lr){2-4}
\cmidrule(lr){5-14}
\cmidrule(l){15-23}
& All & Digit. & Photo.
& Avg. & DE & EN & ES & FR & ID & IT & NL & PT & VI
& Avg. & AR & HI & JP & KO & RU & TH & ZH & ZH-T \\
\midrule

\multicolumn{23}{@{}l}{\textit{General-purpose Vision-Language Models}} \\

\hspace{1.5em}\modelicon{assets/model_logos/table/openai.png}GPT-5.6-Sol
& \underline{86.6} & 90.2 & \textbf{85.6}
& 86.7 & 88.5 & 89.2 & 82.1 & 81.4
& 88.7 & 89.1 & 85.6 & 88.8 & \underline{86.9}
& \underline{84.5} & 88.2 & 88.6 & \underline{76.5} & 84.8
& \textbf{86.2} & \textbf{81.2}
& \underline{86.4} & \underline{84.4} \\

\hspace{1.5em}\modelicon{assets/model_logos/table/gemini.png}Gemini-3-Pro
& 86.4 & 90.4 & \underline{85.1}
& \underline{88.4} & \textbf{91.2} & \underline{90.6}
& \underline{83.4} & \underline{82.7}
& \textbf{91.5} & \underline{91.6} & \underline{87.7}
& \textbf{91.4} & 85.9
& 84.1 & \underline{89.4} & \textbf{90.4}
& 74.8 & 85.5 & 84.9 & 80.6 & 85.1 & 82.1 \\

\hspace{1.5em}\modelicon{assets/model_logos/table/kimi.png}Kimi-K3
& 83.6 & \underline{90.8} & 81.2
& 86.2 & 89.1 & 87.2 & 80.2 & 80.0
& 86.9 & \textbf{92.7} & 86.0 & 88.9 & 84.8
& 80.7 & 77.4 & 77.5 & 74.9 & \textbf{89.9}
& 82.4 & 72.1 & \textbf{89.5} & 81.9 \\

\hspace{1.5em}\modelicon{assets/model_logos/table/qwen.png}Qwen3-VL-Instruct-8b
& 68.3 & 78.4 & 65.0
& 73.6 & 73.7 & 71.4 & 69.3 & 66.2
& 68.5 & 79.1 & 78.3 & 82.2 & 73.4
& 62.5 & 63.1 & 58.4 & 59.9 & 61.9
& 57.9 & 62.0 & 62.6 & 73.8 \\

\hspace{1.5em}\modelicon{assets/model_logos/table/qwen.png}Qwen3.5-Instruct-9B
& 65.7 & 74.8 & 62.7
& 72.5 & 72.8 & 72.0 & 72.0 & 64.4
& 66.2 & 77.6 & 74.5 & 79.1 & 74.0
& 58.2 & 53.4 & 56.2 & 55.7 & 60.3
& 54.7 & 56.7 & 60.8 & 67.5 \\

\hspace{1.5em}\modelicon{assets/model_logos/table/internvl.png}InternVL-3.5-8B
& 42.7 & 59.7 & 37.0
& 53.4 & 39.8 & 64.2 & 47.5 & 42.7
& 53.8 & 60.6 & 52.2 & 63.2 & 57.0
& 30.6 & 8.2 & 9.0 & 45.6 & 30.3
& 26.1 & 10.8 & 55.3 & 59.3 \\

\midrule
\multicolumn{23}{@{}l}{\textit{OCR-specialized Vision-Language Models}} \\

\hspace{1.5em}\modelicon{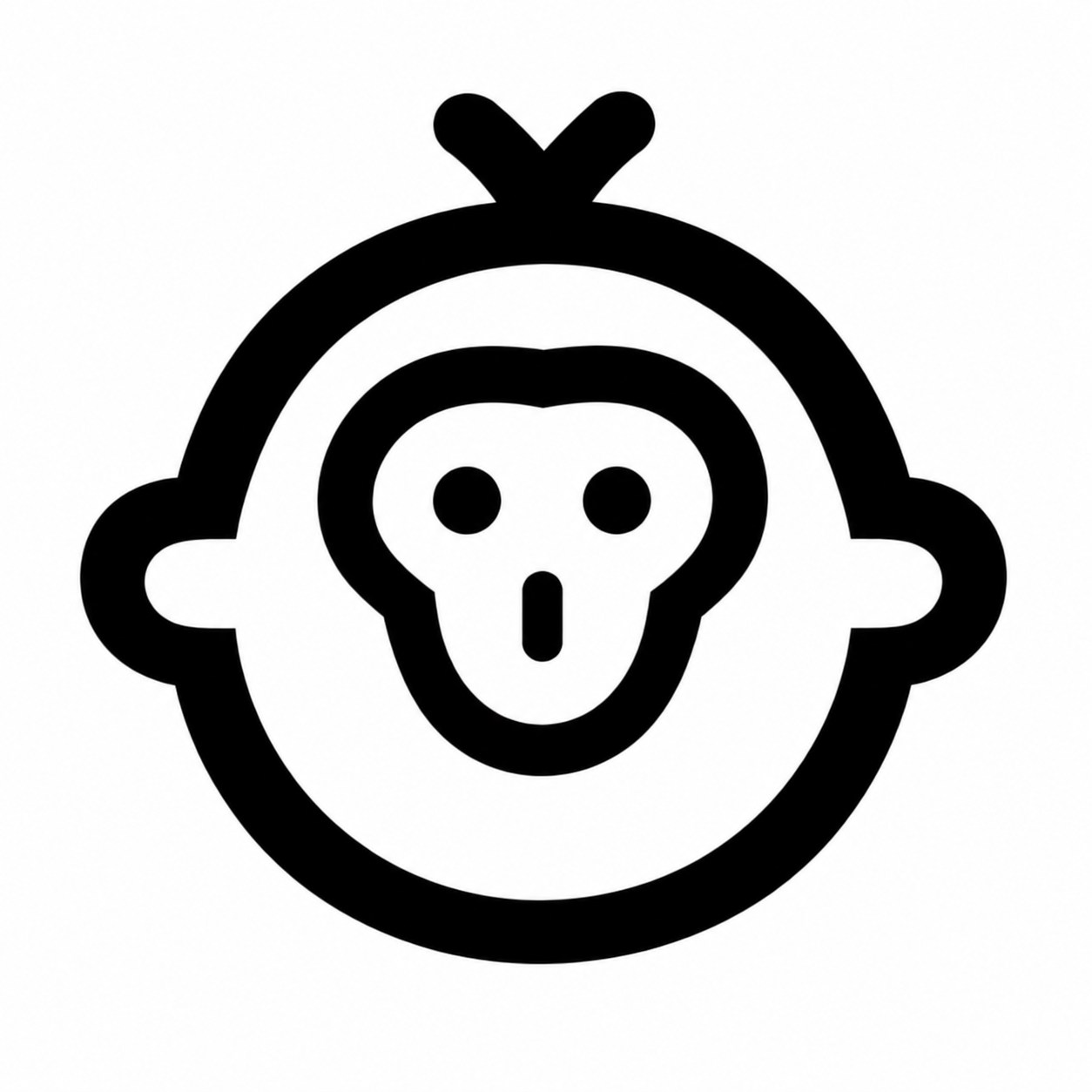}MonkeyOCRv2-B-Parsing
& 83.3 & 88.1 & 81.7
& 84.2 & 87.7 & 84.5 & 75.2 & 78.4
& 86.5 & 88.6 & 86.1 & 87.9 & 83.2
& 82.1 & \textbf{90.7} & 87.2 & 71.9 & \underline{87.6}
& 80.1 & 80.8 & 83.6 & 75.3 \\

\hspace{1.5em}\modelicon{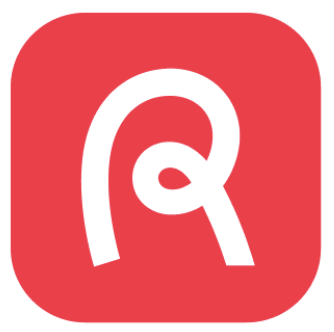}dots.mocr
& 80.5 & 90.5 & 77.2
& 81.7 & 82.6 & 87.4 & 71.3 & 70.1
& 84.5 & 89.3 & 83.2 & 86.8 & 79.9
& 79.2 & 83.3 & 83.6 & 75.0 & 78.7
& 71.2 & 77.9 & 84.6 & 79.6 \\

\hspace{1.5em}\paddleicon PaddleOCR-VL-1.6
& 78.9 & 87.1 & 76.3
& 81.4 & 85.1 & 83.5 & 75.1 & 78.6
& 82.7 & 85.5 & 80.5 & 80.9 & 80.9
& 76.1 & 76.2 & 70.7 & 70.3 & 86.3
& 75.1 & 67.3 & 85.9 & 76.7 \\

\hspace{1.5em}\modelicon{assets/model_logos/table/hunyuan.png}HunyuanOCR-1.5
& 76.8 & 86.2 & 73.6
& 79.7 & 79.6 & 80.4 & 74.2 & 70.0
& 81.5 & 84.5 & 78.4 & 86.4 & 82.4
& 73.5 & 71.8 & 71.6 & 65.5 & 75.7
& 67.4 & 77.7 & 80.8 & 77.2 \\

\hspace{1.5em}\modelicon{assets/model_logos/table/mineru.png}MinerU-2.5-Pro-1.2B
& 71.0 & 86.2 & 66.1
& 74.6 & 78.3 & 79.5 & 63.4 & 67.4
& 78.0 & 79.7 & 72.1 & 78.6 & 74.2
& 67.0 & 56.6 & 72.2 & 59.1 & 77.6
& 62.6 & 61.8 & 76.5 & 69.7 \\

\hspace{1.5em}\paddleicon PaddleOCR-VL
& 69.6 & 87.6 & 63.6
& 72.1 & 78.2 & 79.3 & 62.9 & 66.0
& 77.4 & 78.4 & 67.9 & 72.0 & 66.6
& 66.7 & 65.8 & 68.4 & 59.9 & 77.8
& 56.9 & 57.8 & 78.2 & 68.5 \\

\hspace{1.5em}\modelicon{assets/model_logos/table/ovis.png}OvisOCR2
& 69.3 & 83.7 & 64.5
& 80.2 & 80.4 & \textbf{90.9} & 75.6 & 67.4
& 79.9 & 87.1 & 82.8 & 85.6 & 71.9
& 56.9 & 45.8 & 43.8 & 69.9 & 64.6
& 57.0 & 15.1 & 81.2 & 77.7 \\

\hspace{1.5em}\modelicon{assets/model_logos/table/hunyuan.png}HunyuanOCR
& 68.3 & 80.2 & 64.3
& 72.4 & 75.0 & 73.1 & 63.0 & 66.1
& 69.9 & 80.3 & 61.4 & 81.9 & 80.6
& 63.7 & 68.3 & 73.1 & 55.6 & 68.9
& 52.2 & 60.7 & 66.8 & 64.2 \\

\hspace{1.5em}\modelicon{assets/model_logos/table/glm.png}GLM-OCR
& 67.3 & 77.9 & 63.7
& 78.7 & 82.7 & 84.5 & 75.8 & 76.2
& 79.7 & 82.8 & 80.2 & 77.4 & 69.2
& 54.3 & 21.7 & 39.6 & 65.5 & 61.2
& 64.2 & 27.4 & 78.5 & 76.7 \\

\midrule
\multicolumn{23}{@{}l}{\textit{Pipeline Tools}} \\

\hspace{1.5em}\paddleicon PP-StructureV3
& 45.4 & 56.2 & 41.7
& 59.8 & 60.4 & 68.7 & 54.4 & 49.8
& 69.6 & 68.9 & 55.5 & 58.4 & 52.7
& 28.9 & 1.0 & 7.7 & 56.2 & 15.4
& 7.5 & 11.9 & 72.2 & 59.1 \\

\hspace{1.5em}\modelicon{assets/model_logos/table/mineru.png}MinerU-2.5-pipeline
& 33.5 & 57.6 & 25.4
& 46.5 & 54.3 & 58.3 & 38.4 & 43.6
& 51.9 & 56.5 & 43.9 & 44.0 & 27.6
& 18.7 & 1.2 & 5.3 & 24.5 & 6.8
& 4.2 & 6.4 & 53.9 & 47.2 \\

\midrule
\multicolumn{23}{@{}l}{\textit{\model{} (Ours)}} \\

\hspace{1.5em}\ourmodelicon\model{}-2.7B
& \textbf{87.0} & \textbf{91.0} & \textbf{85.6}
& \textbf{88.8} & \underline{90.5} & 90.2
& \textbf{85.0} & \textbf{84.8} & \underline{90.0}
& 90.8 & \textbf{89.2} & \underline{90.5} & \textbf{88.6}
& \textbf{84.9} & 88.3 & \underline{89.0}
& \textbf{77.2} & 86.8 & \underline{85.4}
& \underline{81.0} & 85.8 & \textbf{85.9} \\

\bottomrule
\end{tabular}%
}
\end{table*}

\subsection{Ablation Study of CGPO}
\label{sec:training_ablation}

The preceding results establish the overall capability of \model. We next compare post-training variants to assess Competence-Guided Policy Optimization. All variants are initialized from the same SFT checkpoint.

The ablation considers five post-training configurations: GRPO alone, OPD alone, GRPO combined with fixed-weight OPD, GRPO combined with cosine-decayed OPD, and the proposed CGPO. These comparisons examine outcome-based policy optimization, on-policy token-level supervision, and sample-wise competence routing alongside a step-dependent distillation schedule.

\paragraph{Step-dependent baseline.}
To test whether a predefined decline in distillation strength accounts for the gains from CGPO, the cosine baseline replaces the sample-dependent coefficient $\alpha(x)$ with
\begin{equation}
    \label{eq:cosine_opd_schedule}
    \alpha_t
    =\alpha_{\min}
    +\frac{\alpha_{\max}-\alpha_{\min}}{2}
    \left(1+\cos\frac{\pi t}{T}\right),
    \qquad 0\leq t\leq T,
\end{equation}
where $t$ is the current optimization step, $T$ is the total number of optimization steps, and $\alpha_{\max}$ and $\alpha_{\min}$ are the initial and final OPD weights, respectively. The joint objective remains $\mathcal{L}_{\mathrm{GRPO}}+\lambda\alpha_t\mathcal{L}_{\mathrm{OPD}}$. At a given step, all samples share the same weight, independent of teacher competence or the teacher--student competence gap.

\paragraph{Component-wise comparison.}

Table~\ref{tab:posttraining_ablation} summarizes the results. GRPO provides direct optimization toward task-level outcome rewards and establishes the contribution of policy optimization beyond SFT. OPD introduces additional token-level supervision along student-generated trajectories. Combining GRPO and OPD with a fixed coefficient further improves performance by jointly exploiting outcome-level and token-level signals.

CGPO achieves the highest scores on all three benchmarks among the evaluated post-training strategies. On CC-OCR, it scores 82.43, improving on the SFT checkpoint by 2.97 points and on fixed-weight GRPO--OPD by 0.25 points. Its routing rule assigns stronger distillation weights when the teacher is estimated to provide reliable, non-redundant information.

Compared with cosine-decayed OPD, CGPO improves OCRBench v2.1 scores by 2.19 and 2.06 points in English and Chinese, respectively, CC-OCR by 0.11 points, and OmniDocBench by 1.45 points. This comparison shows higher benchmark scores for CGPO than for the evaluated step-dependent schedule.

\paragraph{Routing dynamics.}

\begin{figure*}[t]
    \centering
    \begin{minipage}[t]{0.59\textwidth}
        \centering
        \includegraphics[width=\linewidth]{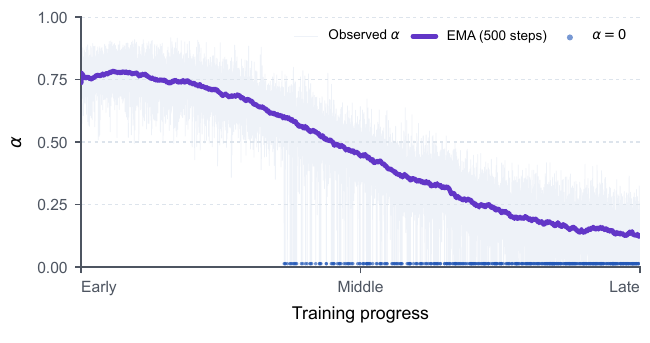}
        \vspace{-1mm}

        {\small\textbf{(a)}}
    \end{minipage}
    \hfill
    \begin{minipage}[t]{0.38\textwidth}
        \centering
        \includegraphics[width=\linewidth]{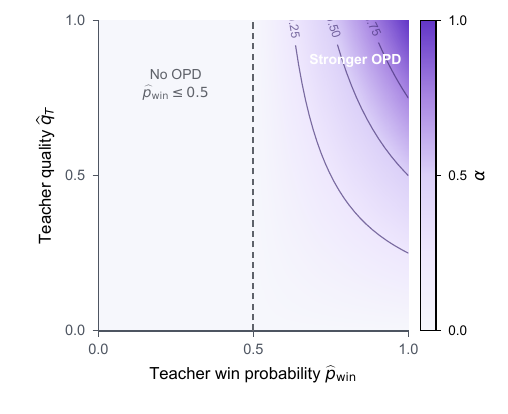}
        \vspace{-1mm}

        {\small\textbf{(b)}}
    \end{minipage}
    \vspace{-1mm}
    \caption{CGPO routing behavior. \textbf{(a)} Observed routing coefficients during CGPO, their exponential moving average, and points with $\alpha=0$. \textbf{(b)} Analytical routing surface induced by $\alpha=\widehat q_T(2\widehat p_{\mathrm{win}}-1)_{+}$. OPD is disabled when $\widehat p_{\mathrm{win}}\leq0.5$ and receives increasing weight as teacher quality and the calibrated probability of teacher superiority increase.}
    \label{fig:cgpo_routing_behavior}
\end{figure*}

Figure~\ref{fig:cgpo_routing_behavior} shows the observed routing coefficient during CGPO. Despite local fluctuations, its smoothed trajectory declines from early to late training, and zero-weight decisions become more frequent toward the later portion of the trace. The analytical surface in panel (b) illustrates how teacher quality and the calibrated probability of teacher superiority jointly determine the OPD weight.

\begin{table*}[t]
    \centering
    \small
    \setlength{\tabcolsep}{2pt}
    \caption{
    Ablation study of Competence-Guided Policy Optimization.
    All variants are initialized from the same SFT checkpoint.
    Cosine decay uses the step-dependent OPD weight in Eq.~\ref{eq:cosine_opd_schedule}, without competence routing.
    }
    \label{tab:posttraining_ablation}
    \resizebox{\textwidth}{!}{
    \begin{tabular}{lcccccc}
        \toprule
        Setting &
        GRPO &
        OPD &
        Competence Routing &
        OCRBench v2.1 (en/zh) &
        CC-OCR &
        OmniDocBench \\
        \midrule

        SFT checkpoint &
        &
        &
        &
        72.08/74.71 &
        79.46 &
        89.72 \\

        + GRPO &
        \checkmark &
        &
        &
        74.63/76.24 &
        81.72 &
        90.54 \\

        + OPD &
        &
        \checkmark &
        &
        73.84/75.77 &
        80.91 &
        90.69 \\

        + GRPO + OPD &
        \checkmark &
        \checkmark &
        &
        75.42/77.08 &
        82.18 &
        91.16 \\

        + GRPO + OPD (cosine) &
        \checkmark &
        \checkmark &
        &
        78.23/76.31 &
        82.32 &
        90.12 \\

        + CGPO &
        \checkmark &
        \checkmark &
        \checkmark &
        \textbf{80.42/78.37} &
        \textbf{82.43} &
        \textbf{91.57} \\

        \bottomrule
    \end{tabular}
    }
\end{table*}

\FloatBarrier

\section{Conclusion}
\label{sec:conclusion}
This report presents \model, a unified OCR foundation model that advances OCR from isolated text recognition toward comprehensive text-centric visual intelligence. By integrating capability-oriented data construction with competence-aware post-training, \model establishes a unified modeling paradigm for diverse OCR scenarios, encompassing scene text understanding, document parsing, information extraction, structured content recovery, multilingual transformation, and OCR-oriented reasoning.

The proposed data construction framework provides large-scale, diverse, and quality-controlled supervision across heterogeneous OCR capabilities by integrating real-world resources, controllable synthesis, and model-assisted annotation. Building upon this foundation, Competence-Guided Policy Optimization (CGPO) further enhances model capabilities through adaptive coordination of outcome-based policy optimization and on-policy distillation. By dynamically allocating teacher supervision according to the evolving competence boundary between teacher and student models, CGPO improves the effectiveness of reinforcement learning and knowledge transfer under heterogeneous OCR tasks.

Extensive evaluations on OCRBench v2.1, CC-OCR, OmniDocBench v1.6, MDPBench, and the in-house KIE benchmark demonstrate that \model achieves strong and balanced performance across text recognition, localization, structured document understanding, multilingual parsing, and real-world information extraction. These results validate the effectiveness of the proposed unified training framework and highlight the feasibility of developing OCR foundation models capable of generalizing across diverse text-containing visual environments while maintaining competitive performance with specialized OCR systems.

Future work will focus on further strengthening fundamental OCR capabilities, such as fine-grained text perception, localization, and recognition, while advancing semantic understanding and reasoning beyond text extraction. Enhancing the generalization and robustness of OCR foundation models in diverse and complex real-world environments remains a key direction toward more reliable text-centric visual intelligence.
\newpage

\bibliographystyle{assets/plainnat}
\bibliography{main}

\newpage
\beginappendix
\section{Contributors}
\label{sec:contri}

\medskip
\noindent\textbf{Core contributors.}
Kaili Huang, Yongshuo Zhang.

\medskip
\noindent\textbf{Contributors.}
Bingtao Fu, Changjiang Jiang, Chenfan Qu, Chenfeng Zhang, Fangming Cui, Gaoyang Zhang, Jiangwei Xie, Jianshu Li, Jing Huang, Jingwen Bai, Mingqi Fang, Tao Fang, Weihong Zhang, Wenbo Du, Xiongfei Bai, Xuekang Zhu, Yinan Xia.
{\let\thefootnote\relax\footnotetext{%
Within each group, names are ordered by first name
in alphabetical order.
}}

\medskip
\noindent\textbf{Data leader.}
Zhenming Wang.

\medskip
\noindent\textbf{Project leaders.}
Jian Liu, Jingjing Liu, Xiang Qi.

\medskip
\noindent\textbf{Project advisor.}
Weiqiang Wang.


    

\section{Implementation and Reproducibility Details}
\label{app:implementation}

This appendix records the implementation details and supplementary evaluation protocols for \model.

\subsection{Model and Input Configuration}

\model is released at three parameter scales. All share the same task interface, data engine, and training settings for SFT and CGPO, including input settings and optimization hyperparameters. Table~\ref{tab:app_model_config} records the base models and input configuration.

\begin{table}[H]
    \centering
    \small
    \caption{Model and input configuration. All variants share the same training settings.}
    \label{tab:app_model_config}
    \begin{tabular*}{0.96\linewidth}{@{\extracolsep{\fill}}lccc}
        \toprule
        Configuration & \modelsmall & \modelbig & \model-2.7B \\
        \midrule
        Base checkpoint & Qwen3.5-2B & Qwen3.5-9B (ModelScope copy) & ViT (1.2B) + Qwen2.5 (1.5B) \\
        Image-area setting (\texttt{MAX\_PIXELS}) & 1,048,576 pixels & 1,048,576 pixels & 1,048,576 pixels \\
        Maximum SFT sequence length & 20,480 tokens & 20,480 tokens & 20,480 tokens \\
        SFT precision & BF16 & BF16 & BF16 \\
        \bottomrule
    \end{tabular*}
\end{table}

The image-area entry records the environment setting in the SFT launch script.


\subsection{Supervised Fine-Tuning Configuration}

Supervised fine-tuning jointly trains the model across text recognition, localization, spotting, key information extraction, document and table parsing, formula recognition, translation, and OCR-centric reasoning.

Both \modelsmall and \modelbig use the same SFT configuration: ms-swift (\texttt{swift sft}) with DeepSpeed ZeRO-2 on 16 nodes with 16 GPU workers per node. We update the language-model parameters in full-tuning mode while freezing the vision encoder and multimodal aligner. Training uses BF16 and the \texttt{flash\_attention\_3} attention implementation. The model and prompt-template types are both \texttt{qwen3\_5}; non-thinking prefixes are enabled, and empty thinking spans are excluded from the loss through \texttt{ignore\_empty\_think}. The training mixture follows the 60M-instance, three-phase schedule in Section~\ref{sec:sft}.

\begin{table}[H]
    \centering
    \small
    \caption{Shared supervised fine-tuning configuration for \modelsmall and \modelbig.}
    \label{tab:app_sft_config}
    \begin{tabular}{lc}
        \toprule
        Configuration & Value \\
        \midrule
        Effective sampled instances & Approximately 60M; three phases of 20M \\
        Maximum SFT sequence length & 20,480 tokens \\
        Task-mixture weights (TP/DS/OR/GA) & $32.5/30.0/22.5/15.0\%$ \\
        GPU workers & 256 ($16$ nodes $\times$ $16$ workers) \\
        Per-device training batch size & 1 \\
        Gradient accumulation steps & 1 \\
        Global training batch size & 256 \\
        Number of epochs & 1 \\
        Peak learning rate and schedule & $10^{-5}$; cosine decay \\
        Warm-up ratio & $5\%$ \\
        Length-based grouping & Enabled \\
        Vision-encoder gradient checkpointing & Disabled \\
        Checkpoint saving interval & 2,000 steps \\
        Checkpoint retention & 4; model weights only \\
        Logging interval & 10 steps \\
        Data-loader workers per process & 12 \\
        Dataset preprocessing processes & 128 \\
        \bottomrule
    \end{tabular}
\end{table}

The global training batch size is $256\times1\times1=256$. Training is configured for one epoch.

\subsection{Competence-Guided Policy Optimization Configuration}

All model variants use the same CGPO configuration. Each starts from its corresponding SFT checkpoint and trains with a learning rate of $10^{-6}$ and Qwen3.5-122B-A10B as the frozen teacher. Teacher competence is estimated from $K=8$ offline generations per sample and cached, whereas student competence is updated from $G=8$ rollouts per group. Both estimates use the same quality-verified reference and task-specific verifier. Table~\ref{tab:app_posttrain_config} summarizes the setup defined in Section~\ref{sec:cgpo}.

Before policy optimization, we fit the teacher-quality and win-probability calibration maps on a fixed split stratified by task and difficulty. Independent $A/B$ partitions of teacher and SFT-student generations separate the router inputs from the calibration targets. For each calibration sample, $|A_T|=|A_S|=8$ matches the competence-estimation budgets used during CGPO, and the independent target partitions contain $|B_T|=|B_S|=8$ generations. Calibration uses task-balanced soft-label binary cross-entropy with positive slopes and repeated independent partitions. The fitted coefficients $a_T,b_T,a_{\Delta},b_{\Delta}$ remain fixed throughout policy optimization.

\begin{table}[H]
    \centering
    \small
    \caption{Shared CGPO setup and calibrated routing for all model variants.}
    \label{tab:app_posttrain_config}
    \renewcommand{\arraystretch}{1.12}
    \begin{tabularx}{\linewidth}{@{}>{\raggedright\arraybackslash}p{0.29\linewidth} >{\raggedright\arraybackslash}X@{}}
        \toprule
        Configuration & Setting \\
        \midrule
        Student initialization & SFT checkpoint \\
        Teacher model & Qwen3.5-122B-A10B; frozen \\
        Learning rate & $10^{-6}$ \\
        Rollout sampling temperature & 0.9 \\
        GRPO clipping & $\epsilon=0.2$; symmetric ratio clipping to $[0.8,1.2]$ \\
        Reference-policy KL penalty & Disabled ($\beta=0$) \\
        Teacher competence $c_T(x)$ & Mean verifier reward over $K=8$ offline generations per sample; cached \\
        Student competence $c_S(x)$ & Mean verifier reward over $G=8$ current student rollouts per group \\
        Calibration split & Fixed training split stratified by task and difficulty \\
        Calibration $A$ budgets & $|A_T|=8$, $|A_S|=8$ per calibration sample \\
        Calibration $B$ budgets & $|B_T|=8$, $|B_S|=8$ per calibration sample; independent of $A$ \\
        Calibration targets & Held-out teacher quality $q_T^B$ and pairwise superiority $p_{\mathrm{win}}^B$; ties contribute $0.5$ \\
        Calibration maps & Positive-slope sigmoid maps for teacher quality and teacher--student competence gap \\
        Calibration fitting & Task-balanced soft-label BCE, $w_x\propto1/N_{\tau(x)}$; repeated independent partitions \\
        Calibration coefficients & Fitted offline; $a_T,b_T,a_{\Delta},b_{\Delta}$ frozen during CGPO \\
        Routing coefficient & $\alpha(x)=\operatorname{sg}[\widehat q_T(x)(2\widehat p_{\mathrm{win}}(x)-1)_+]$ \\
        Zero OPD-weight condition & $\widehat p_{\mathrm{win}}(x)\leq0.5$ \\
        OPD objective & Length-averaged reverse KL, $D_{\mathrm{KL}}(p_\theta\Vert p_T)$, on student-generated prefixes \\
        OPD coefficient $\lambda$ & 1 \\
        Joint optimization & $\mathcal L_{\mathrm{GRPO}}(x)+\lambda\alpha(x)\mathcal L_{\mathrm{OPD}}(x)$ \\
        \bottomrule
    \end{tabularx}
\end{table}

The rollout temperature and clipping coefficient follow the ms-swift GRPO defaults. We disable the reference-policy KL penalty ($\beta=0$); the reverse KL in OPD is retained for teacher distillation. GRPO remains active throughout CGPO, with $\lambda=1$ in the joint objective. The stopped-gradient coefficient $\alpha(x)$ controls the contribution of OPD; when the calibrated teacher superiority is at most $0.5$, the OPD weight is zero. During policy optimization, gradients update the student while the teacher and calibration maps remain frozen.

\section{Task Interfaces and Output Normalization}
\label{app:task_interfaces}

All task families use autoregressive conditional generation while retaining task-appropriate output representations and verifiers. Table~\ref{tab:app_task_interfaces} summarizes the canonical representations used for training and evaluation. Exact serialization grammars, prompt templates, and normalization code will be released with the model and evaluator.

\begin{table}[H]
    \centering
    \scriptsize
    \caption{Task interfaces, output normalization, and scalar verifier rewards.}
    \label{tab:app_task_interfaces}
    \resizebox{\linewidth}{!}{%
    \begin{tabular}{p{2.6cm}p{3.2cm}p{4.2cm}p{2.8cm}}
        \toprule
        Task family & Canonical output & Normalization / validity guard & Scalar reward \\
        \midrule
        Text recognition & Plain-text transcription & Unicode and whitespace normalization; empty or truncated output is invalid & NED similarity \\
        Grounding & One or more bounding boxes & Coordinate-range validation and one-to-one box assignment & IoU \\
        Text spotting & Box--text pairs & Joint schema validation, box matching, and transcription normalization & IoU--NED geometric mean \\
        KIE & Schema-constrained key--value records & Key canonicalization, type-aware value normalization, duplicate-key rejection & Soft field-level F1 \\
        Full-page parsing & Reading-ordered structured text & Element parsing and availability-aware component extraction & NED / CDM / TEDS aggregation \\
        Table extraction & HTML-like table structure & HTML normalization and structural parse validation & TEDS \\
        Formula recognition & LaTeX expression & Syntax normalization followed by rendering validation & CDM \\
        Translation & Translated text & Source-target language validation and deterministic judge schema & LLM-judge score \\
        OCR-centric VQA & Short or open-form answer & Task-specific answer normalization and deterministic judge schema & Exact match or LLM judge \\
        Counting / rule-based reasoning & Answer with optional structured rationale & Output-schema validation and answer recomputation where possible & Rule-based accuracy \\
        \bottomrule
    \end{tabular}}
\end{table}

\subsection{Normalization Order}

Verifier preprocessing follows a fixed order: output parsing, schema validation, task-specific canonicalization, reference-prediction matching, and scalar scoring. Parsing failures and out-of-schema outputs receive zero reward. Canonicalization removes representational differences that do not change the task answer, but does not repair missing fields, hallucinated content, invalid coordinates, or incorrect structure. For all rule-based verifiers, the release will include unit tests covering exact matches, partial matches, malformed outputs, duplicated predictions, and empty outputs.

\subsection{LLM-Judge Controls}

LLM judges are used only when deterministic surface matching does not adequately capture semantic correctness, including translation and selected open-ended VQA tasks. The judge receives the task instruction, reference answer, and model prediction, and returns a constrained continuous score in $[0,1]$. We use Gemini-3.5-Flash-Lite with deterministic decoding. The final release will record the exact model version, prompt, decoding parameters, retry policy, and evaluation date. Judge failures are logged and rerun under the same frozen configuration.

\section{Operational Routing Procedure}
\label{app:routing_procedure}

Before CGPO, we cache teacher competence and fit the calibration maps as described in Section~\ref{sec:cgpo}. During policy optimization, each training sample follows this sequence:
\begin{enumerate}
    \item Retrieve cached teacher competence $c_T(x)$, computed from $K$ offline generations with the task-aligned verifier.
    \item Generate $G$ student rollouts, score them with the same reference and verifier, and compute $c_S(x)$ and $\Delta(x)=c_T(x)-c_S(x)$.
    \item Evaluate the frozen calibration maps $\widehat q_T(x)$ and $\widehat p_{\mathrm{win}}(x)$, then compute
    \begin{equation*}
        \alpha(x)
        =\operatorname{sg}\!\left[
        \widehat q_T(x)
        \left(2\widehat p_{\mathrm{win}}(x)-1\right)_{+}
        \right].
    \end{equation*}
    \item Compute the GRPO loss from the rollout rewards. For samples with a positive OPD weight, evaluate the frozen teacher on the student-generated prefixes and compute the length-averaged reverse-KL loss.
    \item Update the student with $\mathcal L_{\mathrm{GRPO}}(x)+\lambda\alpha(x)\mathcal L_{\mathrm{OPD}}(x)$, keeping the teacher and calibration maps frozen.
\end{enumerate}

When $\widehat p_{\mathrm{win}}(x)\leq0.5$, $\alpha(x)=0$ and the joint objective reduces to the GRPO term. Otherwise, the OPD contribution varies continuously with the calibrated teacher quality and superiority estimate.

\section{OCRBench v2.1 Evaluation Protocol}
\label{app:ocrbenchv21}

To comprehensively evaluate OCR capabilities, we adopt OCRBench v2~\citep{ocrbenchv2} as one of our primary evaluation benchmarks. OCRBench v2 is a large-scale bilingual, text-centric benchmark with 10,000 human-verified question-answer pairs spanning diverse visual scenarios. Throughout this report, subtask counts follow the evaluation export: 30 language-specific subtasks, comprising 21 English and 9 Chinese subtasks. These are aggregated into 13 language-specific task groups, comprising 8 English and 5 Chinese groups.

To further improve evaluation reliability and consistency, we conduct a systematic audit of both benchmark annotations and the official evaluation protocol. Our analysis identifies two aspects that may affect evaluation robustness: (i) annotation noise and format inconsistencies in reference answers, and (ii) evaluation criteria that may not fully align with the intended semantic objectives of certain OCR tasks.

Based on this analysis, we introduce \textbf{OCRBench v2.1}, a refined version of the benchmark that retains the original 10,000 samples and uses the same 30-subtask reporting scheme while incorporating manually verified annotation corrections and revised evaluation procedures. The refinement focuses on improving reference quality and metric consistency, without modifying the benchmark composition, task distribution, or evaluation scope.

All models reported in this work are re-evaluated under the OCRBench v2.1 protocol to ensure fair comparison. Detailed descriptions of annotation refinement and metric revisions are provided in Sections~\ref{sec:v21-annotation} and~\ref{sec:v21-metric}, respectively.

\subsection{Annotation Cleaning}
\label{sec:v21-annotation}

\paragraph{Motivation for annotation refinement.}
High-quality reference annotations are essential for reliable OCR evaluation, particularly for tasks involving structured prediction and semantic understanding. Under the original evaluation protocol, certain annotation issues may have limited impact because some metrics emphasize structural similarity or partial matching. However, under stricter content-aware evaluation, inaccurate references or inconsistent output formats can introduce additional penalties unrelated to the actual model capability.

In addition, ambiguous task instructions and inconsistent answer serialization may affect the stability of automated evaluation procedures, especially for structured outputs such as JSON, HTML, and field-value extraction. Therefore, improving annotation consistency is a necessary step toward more reliable and reproducible benchmark evaluation.

\paragraph{Scope of refinement.}
We conduct a comprehensive audit of all 10,000 samples, covering questions, answers, evaluation configurations, bounding boxes, and reference contents. The audit results in revisions to \textbf{2,449 samples (24.5\%)}. No samples are added or removed, and the original indices, task categories, bounding boxes, and evaluation configurations remain unchanged. All modifications are restricted to the \texttt{question} and \texttt{answer} fields and are manually inspected for verification.

The annotation revisions can be categorized into three types, summarized in Table~\ref{tab:v21-annotation}: (A) format normalization, which improves structural consistency without changing the underlying semantics; (B) reference correction, which fixes inaccurate or incomplete ground-truth answers; and (C) instruction refinement, which reduces ambiguity in task descriptions and output requirements.

\begin{table}[t]
\centering
\small
\caption{Annotation revisions in OCRBench v2.1. Type A normalizes answer serialization without changing semantics; Type B corrects genuinely wrong references; Type C rewrites extraction instructions to remove ambiguity. ``\#mod'' counts revised samples per type; a sample modified in both \texttt{question} and \texttt{answer} is counted once in each type, so the column sums slightly exceed the 2,449 unique revised samples.}
\label{tab:v21-annotation}
\begin{tabular}{llrc}
\toprule
\textbf{Type} & \textbf{Affected subtasks} & \textbf{\#mod} & \textbf{Field} \\
\midrule
\multirow{4}{*}{A. Format} & chart parsing (en)        & 400 & answer \\
                           & key information mapping (en) & 300 & answer \\
                           & table parsing (en / cn)   & 700 & answer \\
                           & text grounding (en)       & 150 & answer \\
\cmidrule(lr){1-4}
B. Content   & cognition / reasoning / diagram / science / math QA, & 626 & answer \\
             & APP agent, text translation (cn), KIE (en \& cn) & & \\
\cmidrule(lr){1-4}
C. Question  & key information extraction (en: 400; cn: 184) & 584 & question \\
\midrule
\textbf{Total} & \textbf{16 of 30 subtasks} & \textbf{2,449} & \\
\bottomrule
\end{tabular}
\end{table}

\paragraph{Type A: format normalization (1,550 samples).}
We remove redundant formatting noise and standardize structured annotations to improve evaluation consistency without changing their semantic content. For table parsing, we eliminate scattered CSS styles, wrapper elements, and non-essential attributes, retaining only the basic HTML tags required to represent table structures. For other structured tasks, chart parsing and key-information-mapping references are converted from Python-dict-style representations into valid JSON strings, and text-grounding coordinates are uniformly serialized to ensure consistent parsing and matching during evaluation.

\paragraph{Type B: reference correction (626 samples).}
We correct inaccurate or incomplete reference annotations identified during manual inspection. These corrections mainly occur in multiple-choice reasoning tasks, key information extraction, and text translation.

Typical corrections include erroneous option labels in mathematical reasoning questions, swapped or inaccurate field values in KIE annotations, OCR-related text corrections, unified formatting of currency symbols, and revised translation references. For cases where multiple answers are semantically acceptable, all valid alternatives are explicitly enumerated and supported by the v2.1 evaluation protocol (see Section~\ref{sec:v21-metric}).

\paragraph{Type C: instruction refinement (584 samples).}
We refine task instructions for key information extraction, where ambiguous extraction requirements may introduce uncertainty beyond the actual recognition and extraction capability.

For English KIE samples, we rewrite all 400 questions into a unified template that explicitly specifies the target fields to extract. For Chinese KIE samples, we append explicit JSON-format requirements and normalize field naming conventions for 184 questions. These refinements ensure that extraction metrics primarily reflect the model's ability to identify and organize information from visual inputs, rather than its interpretation of underspecified output formats.

\subsection{Metric Revisions}
\label{sec:v21-metric}

To improve the reliability and task alignment of evaluation, we revise the metrics for several OCRBench v2 subtasks where the original criteria may not fully reflect the intended capability being measured. Table~\ref{tab:v21-metric} summarizes the differences between the original v2 protocol and the v2.1 evaluation scheme.

All remaining subtasks retain the original evaluation metrics, including exact matching / ANLS for VQA~\citep{biten2019stvqa}, TEDS for tables~\citep{zhong2020pubtabnet}, IoU-based metrics for grounding and spotting, F1 for information extraction, and normalized L1 for counting. The official aggregation strategy, including the 13 task groups and English/Chinese overall scores, is also preserved. Therefore, the reported differences arise only from the refined annotations and revised evaluation criteria described below.

\begin{table*}[t]
\centering
\small
\setlength{\tabcolsep}{6pt}
\caption{Metric revisions from OCRBench v2 to v2.1. Only the listed subtasks are revised; other subtasks follow the official evaluation protocol.}
\label{tab:v21-metric}
\begin{tabular*}{\textwidth}{@{\extracolsep{\fill}}lccc@{}}
\toprule
\textbf{Subtask} & \textbf{OCRBench v2} & \textbf{OCRBench v2.1}\\
\midrule

Formula recognition
& Substring matching (\texttt{gt in pred})
& Render-based CDM with token-level $F_1$ \\

Document parsing
& STEDS (structure-only similarity)
& Content-aware md2md evaluation \\

Fine-grained OCR
& BLEU, METEOR, token-$F_1$, and edit-distance average
& Normalized edit distance \\

Full-page OCR
& BLEU, METEOR, token-$F_1$, and edit-distance average
& Normalized edit distance \\

Text translation
& $n$-gram metric average
& LLM-based semantic evaluation \\

Reasoning VQA
& Single-answer exact matching
& Multiple acceptable references \\

\bottomrule
\end{tabular*}
\end{table*}

\paragraph{(1) Formula recognition: from substring matching to render-then-match CDM.}
The original scorer determines correctness based on whether the reference formula appears as a substring of the prediction after removing whitespace. This criterion may be insufficient for formula understanding, as it can accept incomplete semantic matches while rejecting visually equivalent formulations with different LaTeX syntax.

We replace it with CDM~\citep{wang2024cdm}, where both reference and prediction formulas are rendered and compared at the visual token level. The metric performs token matching with geometric verification and computes an $F_1$ score:
\begin{equation}
F_1^{\mathrm{CDM}} = \frac{2PR}{P+R}, \qquad
P = \frac{|M|}{|T_p|}, \quad R = \frac{|M|}{|T_g|},
\end{equation}
where $T_p$ and $T_g$ denote the visual tokens of the prediction and reference, and $M$ represents the verified token matches.

This evaluation is more robust to common LaTeX syntax variations while remaining sensitive to the visual correctness of mathematical expressions.

\paragraph{(2) Document parsing: from skeleton similarity to content-aware md2md evaluation.}
The original implementation evaluates document parsing mainly through structural similarity. While this captures high-level organization, it provides limited assessment of whether the recovered textual content is correct.

We adopt the md2md protocol from OmniDocBench~\citep{ouyang2025omnidocbench}, where predicted and reference Markdown documents are decomposed into typed elements, including text, formulas, and tables. Each element is evaluated using an appropriate metric, and the final score is computed as:
\begin{equation}
S_{\text{md2md}} = \frac{1}{|\mathcal{C}|} \sum_{c \in \mathcal{C}} \mathrm{Acc}(c),
\qquad
\mathrm{Acc}(c) =
\begin{cases}
S_{\mathrm{ED}}       & c = \text{text},\\
F_1^{\mathrm{CDM}}    & c = \text{formula},\\
\mathrm{TEDS}         & c = \text{table}.
\end{cases}
\end{equation}
This makes document parsing scored at the character, formula-rendering, and table-structure levels simultaneously.

\paragraph{(3) Text recognition and full-page OCR: normalized edit distance.}
The original protocol combines multiple metrics, including BLEU, METEOR, token-level $F_1$, and edit distance. However, for OCR recognition tasks with relatively short outputs, these metrics may provide inconsistent signals and lack a unified interpretation.

We replace this aggregation with normalized edit distance:
\begin{equation}
S_{\mathrm{ED}}(p, g) = \max\!\left(0,\; 1 - \frac{\mathrm{EditDist}(p, g)}{\max(|p|,\, |g|)}\right),
\end{equation}
which directly measures character-level recognition accuracy. The metric is deterministic, monotonic, and closely aligned with the objective of OCR text transcription.

\paragraph{(4) Text translation: LLM-based semantic evaluation.}
LLM-based assessment has been studied for both reference-based and reference-free translation evaluation~\citep{kocmi2023gemba}.
Translation quality is primarily determined by semantic preservation rather than surface-level word overlap. Therefore, $n$-gram based metrics may not fully capture acceptable paraphrases, synonym substitutions, or different word orders.

We evaluate translation quality using an LLM-based judge, \textbf{Gemini-3.5-Flash-Lite}, with the fixed evaluation prompt shown in Figure~\ref{fig:v21-translation-prompt}. The judge evaluates semantic equivalence while considering translation completeness, factual consistency, and target-language correctness. The output score in $[0,1]$ is directly used as the sample-level evaluation score. For samples with multiple acceptable references, the maximum score over references is adopted.

\begin{figure}[t]
\centering
\fbox{\begin{minipage}{0.94\linewidth}
\footnotesize\ttfamily
You are a translation evaluator for OCRBench v2. The task may be Chinese-to-English or English-to-Chinese. Compare the model's translated text with the reference translation and judge whether they express the same meaning in the intended target language.\\[2pt]
Evaluation principles:\\
- Focus on semantic equivalence, not word-by-word matching.\\
- Natural paraphrases, synonym substitutions, and word-order differences should receive high scores if the meaning is preserved.\\
- Do not penalize harmless formatting or punctuation differences.\\
- Penalize mistranslation, missing key information, wrong entities/numbers/negation/ relations, or using the wrong target language.\\
- If the model output contains extra explanation, score based on whether the reference meaning is still complete and not contradicted.\\[2pt]
Model translation:\\
\{predict\}\\[2pt]
Reference translation:\\
\{answer\}\\[2pt]
Please give a score between 0 and 1:\\
- 1.0: Semantically equivalent, fluent, and complete\\
- 0.7--0.9: Mostly equivalent with only minor issues\\
- 0.4--0.6: Partially correct but with obvious omissions or mistranslations\\
- 0.1--0.3: Only a few fragments are relevant; overall meaning is wrong\\
- 0.0: Completely wrong, unrelated, empty, or unreadable\\[2pt]
Output ONLY a single number between 0 and 1 (e.g. 0.8). Do not output anything else.
\end{minipage}}
\caption{The judge prompt used for the text-translation subtask in OCRBench v2.1. \texttt{\{predict\}} and \texttt{\{answer\}} are replaced by the model output and one reference translation, respectively.}
\label{fig:v21-translation-prompt}
\end{figure}

\paragraph{(5) Reasoning VQA (cn): multi-answer support.}
Some reasoning VQA samples may contain multiple semantically acceptable answers. The original evaluation protocol assumes a single reference answer, which may introduce unnecessary penalties for such cases.

The v2.1 scorer supports multiple valid references by accepting a prediction if it matches any enumerated answer. For single-answer samples, the evaluation behavior remains unchanged. This modification improves evaluation consistency for reasoning tasks with answer ambiguity.

\subsection{Comparability and Reproducibility}
\label{sec:v21-comparability}

The above refinements introduce two important considerations for evaluation interpretation.

First, OCRBench v2.1 scores should not be directly compared with the original OCRBench v2 leaderboard results, since the protocol incorporates annotation corrections and revised metrics for several subtasks. All OCRBench v2.1 results reported in this work are obtained under the same refined annotations and scoring procedures.

Second, the v2.1 evaluation pipeline is designed to be deterministic and reproducible. All metrics are computed offline with fixed procedures, with the exception of the LLM-based translation evaluator, which uses Gemini-3.5-Flash-Lite with a fixed evaluation prompt shown in Figure~\ref{fig:v21-translation-prompt}. The refined annotations and evaluation code are released together with this report to facilitate reproducibility.

\section{Task Examples}
\label{app:qualitative}

\begin{CJK}{UTF8}{gbsn}

\begingroup
\newcolumntype{T}{>{\raggedright\arraybackslash}m{2.4cm}}
\newcolumntype{Q}{>{\raggedright\arraybackslash}m{4.5cm}}
\newcolumntype{A}{>{\raggedright\arraybackslash}m{4.5cm}}
\newcolumntype{I}{>{\centering\arraybackslash}m{2.8cm}}

\newcommand{\tableimage}[1]{%
    \includegraphics[width=2.6cm]{#1}%
}

\renewcommand{\arraystretch}{1.35}
\small

\setlength{\tabcolsep}{4pt}

\setlength{\LTleft}{\fill}
\setlength{\LTright}{\fill}

\begin{longtable}{| T | Q | A | I |}

        \caption{Data examples.}
        \label{app:fine_grained_qa} \\
        \hline
        \textbf{Type} &
        \textbf{Question} &
        \textbf{Answer} &
        \textbf{Image} \\
        \hline
        \endfirsthead

        \multicolumn{4}{c}{\tablename~\thetable{} (continued)} \\
        \hline
        \textbf{Type} &
        \textbf{Question} &
        \textbf{Answer} &
        \textbf{Image} \\
        \hline
        \endhead

        \hline
        \multicolumn{4}{r}{Continued on next page} \\
        \endfoot

        \hline
        \endlastfoot

        Spotting &
        Spotting all the text in the image with line-level. Output the normalized coordinates of the left-top and right-bottom corners of the bounding box and the text content. The coordinates should be normalized ranging from 0 to 1000 by the image width and height.\newline
        Your answer should be in the following format:\newline
        [(x1, y1, x2, y2, text content), (x1, y1, x2, y2, text content)...] &
        [(148, 144, 907, 269, "北京北硬硬质合金有限公司"), (392, 234, 707, 334, "经营部"), (100, 606, 154, 621, "北河沿大街"), (96, 619, 161, 643, "11号")] &
        \tableimage{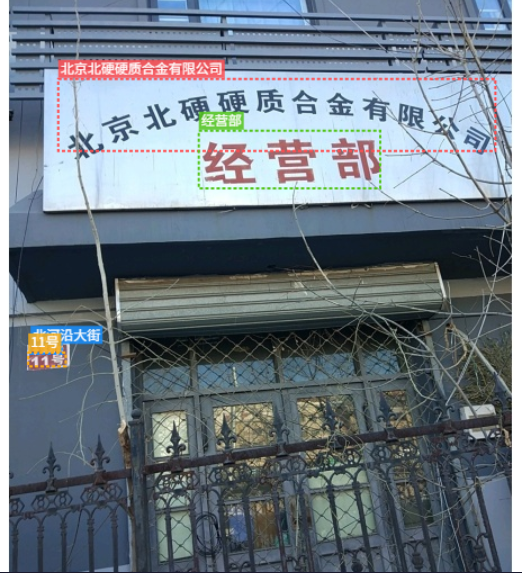} \\
        \hline


        \makecell[l]{Fine-grained\\OCR} &
        Recognize the text within the [295, 390, 352, 432] of the image. The coordinates have been normalized ranging from 0 to 1000 by the image width and height. &
        679 &
        \tableimage{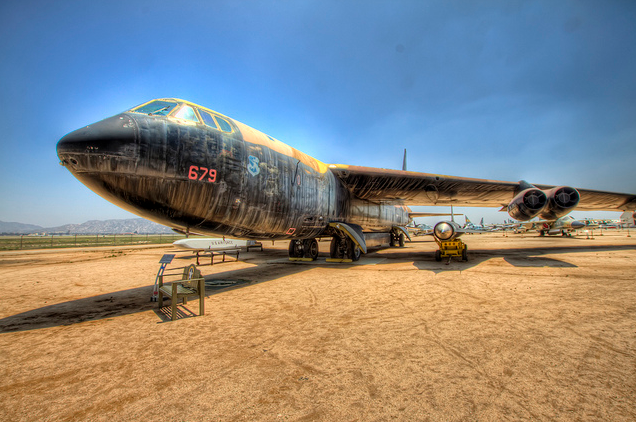} \\
        \hline


        \makecell[l]{Text\\Recognition} &
        what is written in the image? &
        JCPenney &
        \tableimage{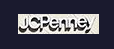} \\
        \hline


        Text VQA &
        what are the letters on the tail section of the plane? Output the answer with 'answer' and 'bbox'. 'bbox' refers to the bounding box position of the 'answer' content in the image. The output format is "answer:gt, bbox:(x1,y1,x2\hspace{0pt},y2)", where the bbox is the coordinates of the top-left corner and the bottom-right corners. The 'bbox' should be normalized coordinates ranging from 0 to 1000 by the image width and height. Your answer should be in the JSON format:
        \{ "answer": "..", \# The answer; "bbox": "(x1,y1,x2\hspace{0pt},y2)" \# The bounding box position of the 'answer' \} &
        \{"answer": "EC", "bbox": "(240,530,295,\hspace{0pt}576)"\} &
        \tableimage{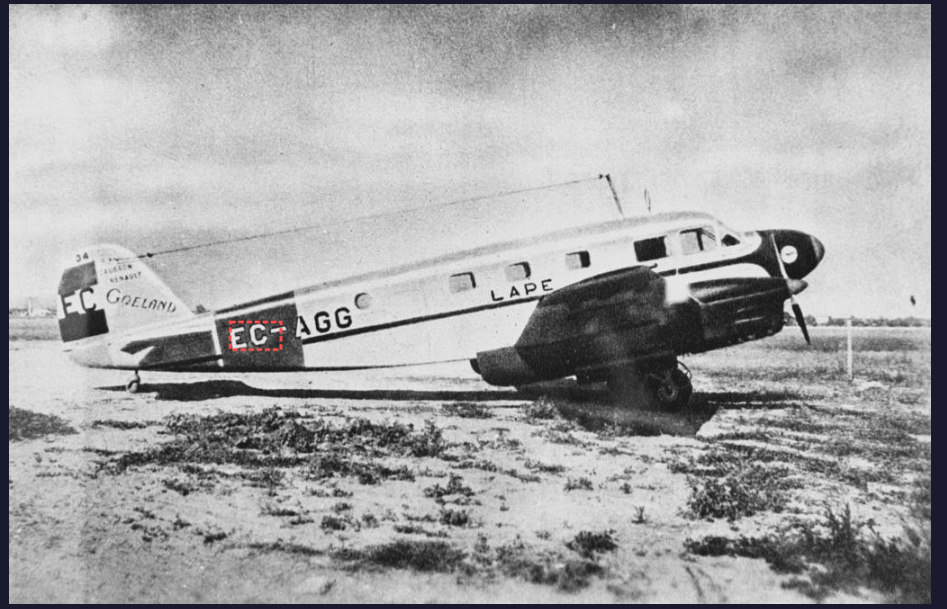} \\
        \hline


        \makecell[l]{Key Info\\Extraction} &
        从图中提取: 样品特征和状态, 主检, 检验类别, 审核, 样品名称, 并按json格式返回 &
        \{"样品特征和状态": "外观正常", "主检": "\#\#\#", "检验类别": "委托送样检验", "审核": "\#\#\#", "样品名称": "玉米芯颗粒粉"\} &
        \tableimage{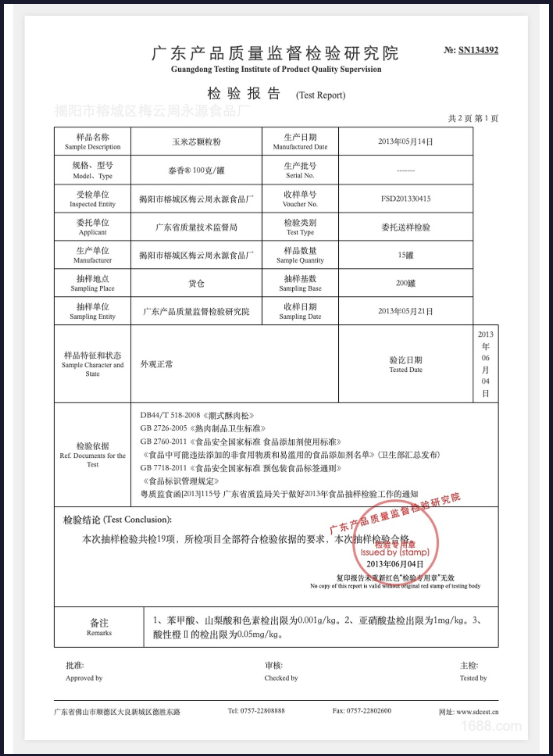} \\
        \hline


        \makecell[l]{Formula\\Recognition} &
        Please write out the expression of the formula in the image using LaTeX format. &
        \$\textbackslash{}mathbb\{E\}\_\{x \textbackslash{}sim p\_t\}\textbackslash{}left[
        \textbackslash{}log \textbackslash{}left(
        1 + \textbackslash{}eta\_t m(\textbackslash{}lambda\_0,x)
        \textbackslash{}right)
        \textbackslash{}right]
        = (1 - p) \textbackslash{}log(1 - \textbackslash{}eta\_t \textbackslash{}alpha)
        + p \textbackslash{}log(1 + \textbackslash{}eta\_t (1-\textbackslash{}alpha))\$ &
        \tableimage{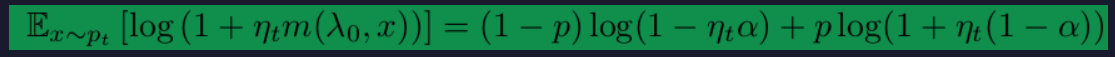} \\
        \hline


        \makecell[l]{Table Parsing\\(region)} &
        Describe the specific HTML-formatted\hspace{0pt} structure of the table in the region [117, 419, 881, 656] of the image. The coordinates have been normalized ranging from 0 to 1000 by the image width and height. &
        \textless{}table\textgreater{}
        \textless{}tr\textgreater{}
        \textless{}td\textgreater{}Site number\textless{}/td\textgreater{}
        \textless{}td\textgreater{}River\textless{}/td\textgreater{}
        \textless{}td\textgreater{}Restored / Channelised\textless{}/td\textgreater{}
        \textless{}td\textgreater{}Basal area density ( \$ m\^{}\{2\} \$ ha \$ \^{}\{-1\} \$)\textless{}/td\textgreater{}
        \textless{}td\textgreater{}\% Picea abies (by \$ m\^{}\{2\} \$)\textless{}/td\textgreater{}
        \textless{}td\textgreater{}\% Pinus sylvestris (by \$ m\^{}\{2\} \$)\textless{}/td\textgreater{}
        \textless{}/tr\textgreater{}......\textless{}/table\textgreater{} &
        \tableimage{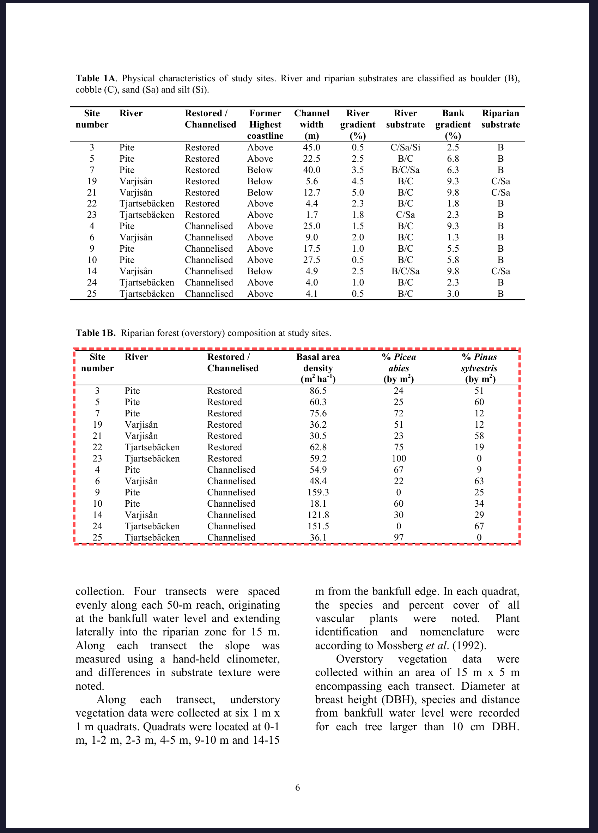} \\
        \hline


        Table Parsing &
        Reconstruct the table in the presented picture as an HTML-style table in text. &
        \textless{}table\textgreater{}
        \textless{}tr\textgreater{}
        \textless{}td\textgreater{}Type of delay\textless{}/td\textgreater{}
        \textless{}td\textgreater{}Mean\textless{}/td\textgreater{}
        \textless{}td\textgreater{}SD\textless{}/td\textgreater{}
        \textless{}td\textgreater{}Median\textless{}/td\textgreater{}
        \textless{}td\textgreater{}IQR\textless{}/td\textgreater{}
        \textless{}td\textgreater{}Min\textless{}/td\textgreater{}
        \textless{}td\textgreater{}Max\textless{}/td\textgreater{}
        \textless{}/tr\textgreater{}......\textless{}/table\textgreater{} &
        \tableimage{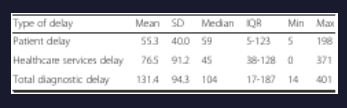} \\
        \hline


        \makecell[l]{Chart\\Understanding} &
        Is the value of Favorable 38 in 2015? &
        Yes &
        \tableimage{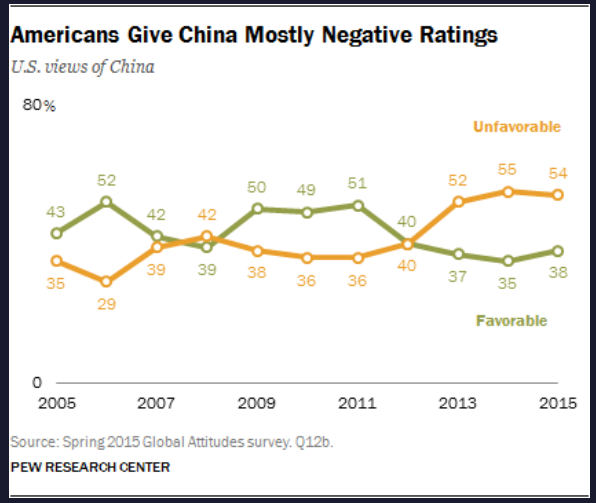} \\
        \hline


        Full-page OCR &
        Read and output all the text content from this image. &
        第22卷第6期\newline
        2020年12月\newline
        辽宁工业大学学报(社会科学版)\newline
        Journal of Liaoning University of Technology(Social Science Edition)\newline
        Vol.22,No.6\newline
        Dec.2020\newline
        DOI: 10.15916/j.issn1674-327X.2020.06.009\newline
        区块链技术在企业财务舞弊治理中的应用\newline
        ...... 图1 虚增存货数总额\newline
        (C)1994-2020 China Academic Journal Electronic Publishing House. All rights reserved. http://www.cnki.net &
        \tableimage{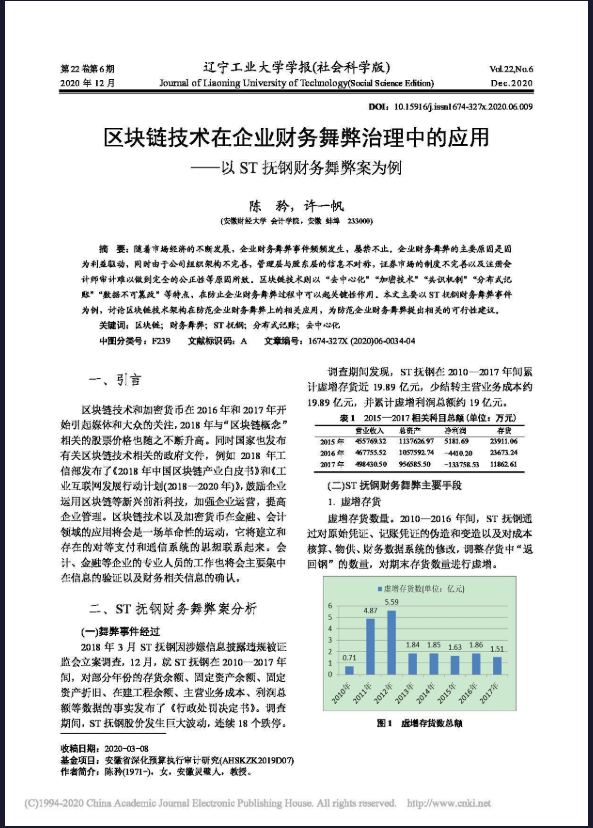} \\
        \hline


        \makecell[l]{Document\\Parsing} &
        \textit{[System prompt: convert document page images to Markdown. Transcribe all visible text preserving reading order and structure; convert formulas to LaTeX; convert tables to HTML; do not describe figures; return only the converted Markdown.]} &
        2852\newline
        自然资源学报\newline
        35卷\newline
        式中： \$ P\_\{D,U\} \$ 为要素D对乡村旅游的影响力；n 为研究区样本数（个），n=55； \$ \textbackslash{}sigma\^{}\{2\} \$ 和 \$ \textbackslash{}sigma\_\{i\}\^{}\{2\} \$ 分别为所有区域与次级区域的方差；m 为次级区域数（个）。 \$ \textbackslash{}sigma\^{}\{2\} \textbackslash{}neq 0 \$，模型成立。 \$ P\_\{D,U\} \$ 的取值区间为 [0, 1]， \$ P\_\{D,U\} \$ 的值越大，说明要素D对乡村旅游热度的影响力就越强，反之则越弱。\newline
        ...... 图2 2009---2017年江苏省乡村旅游热度的月统计图\newline
        Fig. 2 Monthly count of rural tourism popularity in Jiangsu province from 2009 to 2017\newline
        (C)1994-2020 China Academic Journal Electronic Publishing House. All rights reserved. http://www.cnki.net &
        \tableimage{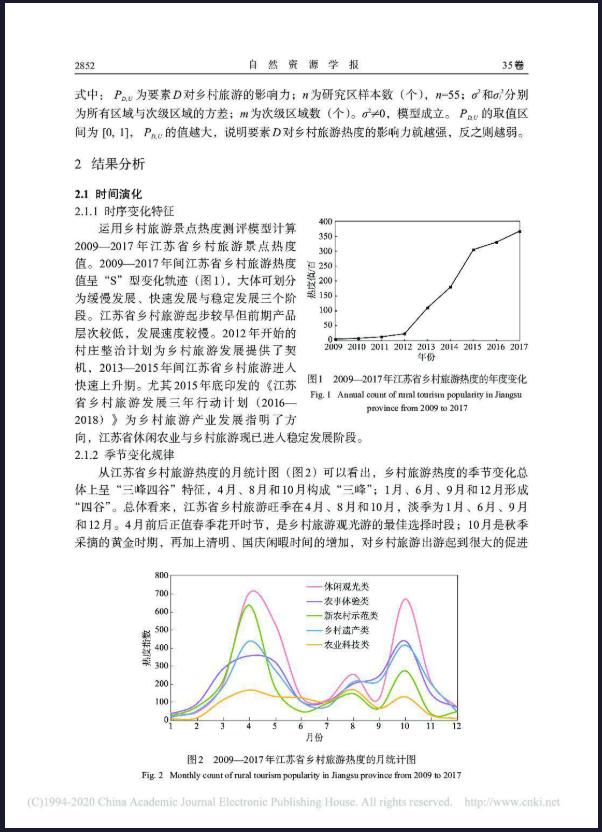} \\
        \hline


        OCR-VQA &
        What is the title of this book? &
        Underground &
        \tableimage{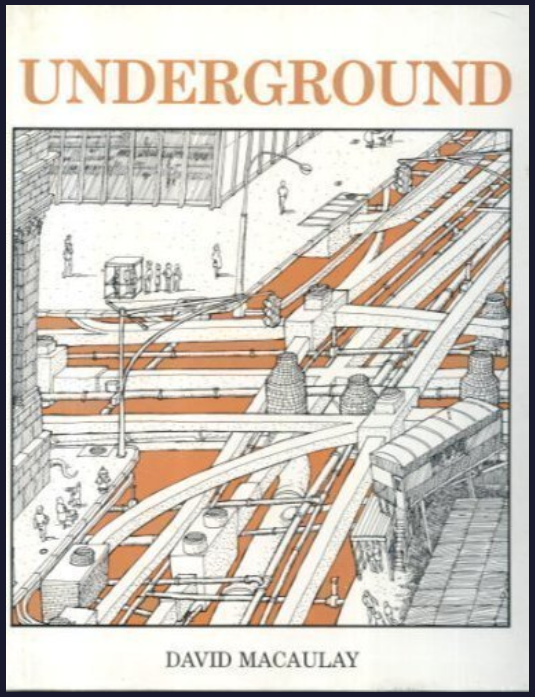} \\
        \hline


        \makecell[l]{Education \&\\Science QA} &
        After the Akkadian Empire ended, many cities and empires tried to control Mesopotamia. Around the 1790s BCE, which empire started controlling Mesopotamia?\newline
        A. the Elamite Empire\newline
        B. the Neo-Sumerian Empire\newline
        C. the Babylonian Empire\newline
        D. the Akkadian Empire &
        C &
        \tableimage{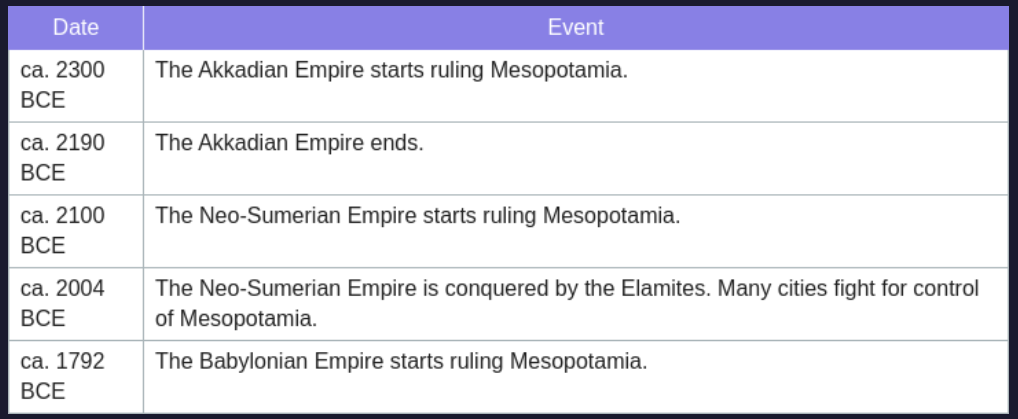} \\
        \hline


        \makecell[l]{Scene Text\\Translation} &
        What is the Chinese translation of the text inside the rectangle defined by [357, 137, 659, 309] in this image? The coordinates have been normalized ranging from 0 to 1000 by the image width and height. Please provide the translated content directly. &
        爱 &
        \tableimage{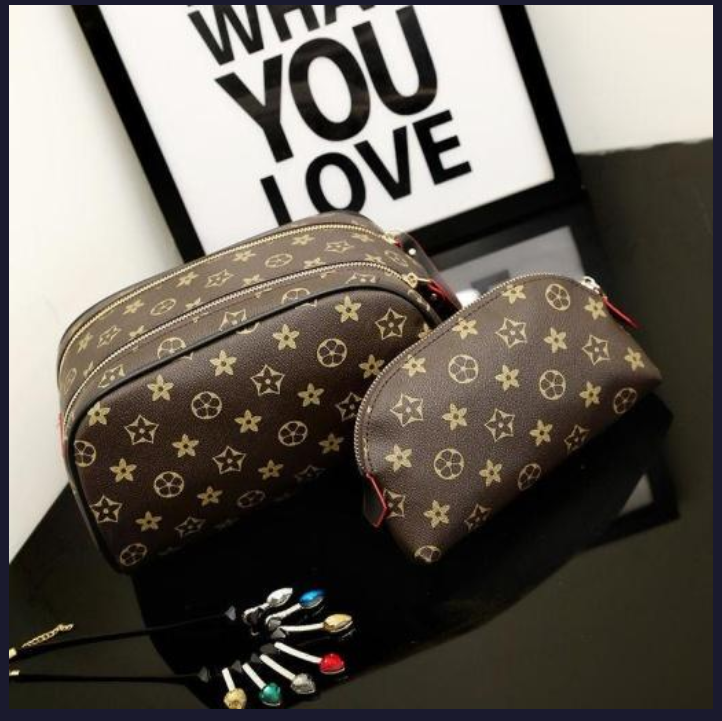} \\
        \hline


        APP Agent &
        How many messages are unread? &
        There is 1 unread message. &
        \tableimage{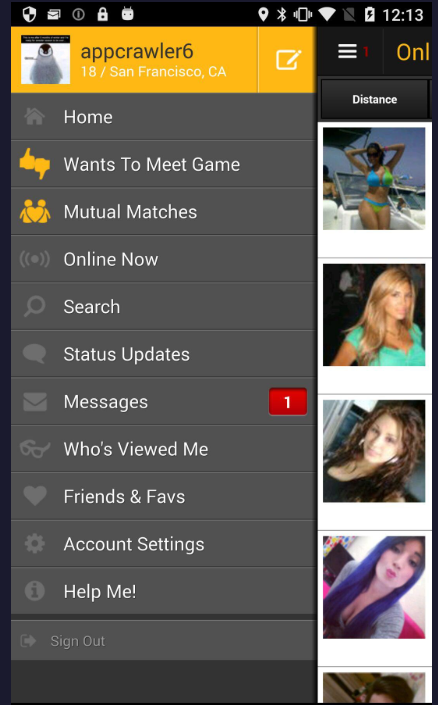} \\
        \hline


        \makecell[l]{Text\\Counting} &
        How many times does the character 'F' appear in the picture? Please output the exact number without any additional explanation. &
        2 &
        \tableimage{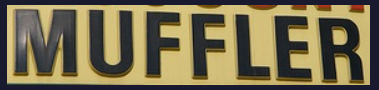} \\
        \hline


        ASCII Art QA &
        What is depicted in this ASCII art image?\newline
        A. dancing\newline
        B. eating\newline
        C. running\newline
        D. singing &
        A. dancing &
        \tableimage{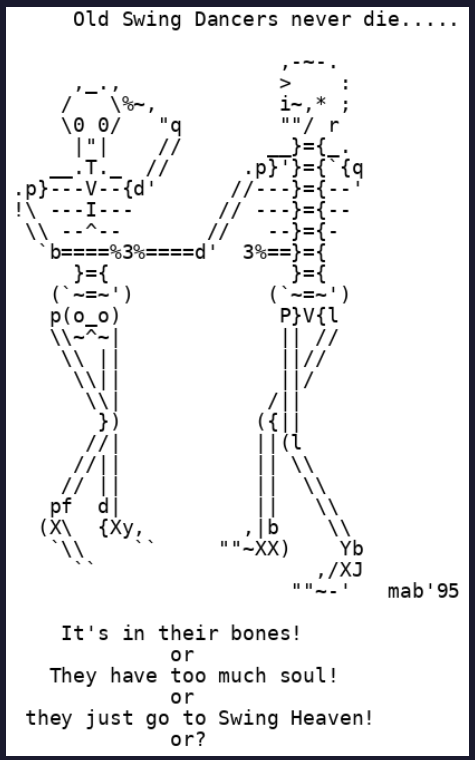} \\
        \hline


        \makecell[l]{Chart Parsing\\} &
        Convert the key information in the chart into a nested python dict. &
        \{'title': 'title', 'source': 'Data source: ', 'x\_title': 'xaxis\_label', 'y\_title': ['yaxis\_label'], 'values': \{'Royal Blue': \{'xaxis\_label': 'Royal Blue', 'yaxis\_label': 74.5866897989292\}, 'Pale Green': \{'xaxis\_label': 'Pale Green', 'yaxis\_label': 88.21400944286917\}, 'Dark Red': \{'xaxis\_label': 'Dark Red', 'yaxis\_label': 7.675044699564133\}, 'Light Green': \{'xaxis\_label': 'Light Green', 'yaxis\_label': 39.7387773448238\}, 'Dark Salmon': \{'xaxis\_label': 'Dark Salmon', 'yaxis\_label': 17.0608316600648\}\}\} &
        \tableimage{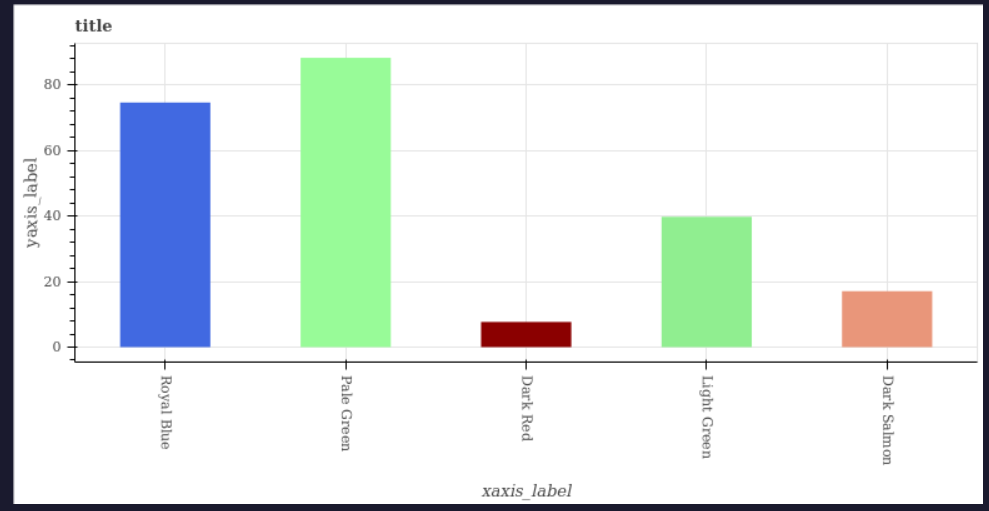} \\
        \hline


        \makecell[l]{Chart Parsing\\} &
        Convert the key information in the chart into a nested python dict. &
        \{'title': 'Distance Education Enrollment by Country', 'source': 'Data Source: International Educational Statistics, Research Reports', 'x\_title': 'None', 'y\_title': ['None', '\%'], 'values': \{'Enrollment (Million Students)': \{'2015': '50', '2016': '55', '2017': '60', '2018': '65', '2019': '70', '2020': '75'\}, 'YOY Growth (\%)': \{'2015': 'None', '2016': '5.0\%', '2017': '6.0\%', '2018': '8.0\%', '2019': '10.0\%', '2020': '12.0\%'\}\}\} &
        \tableimage{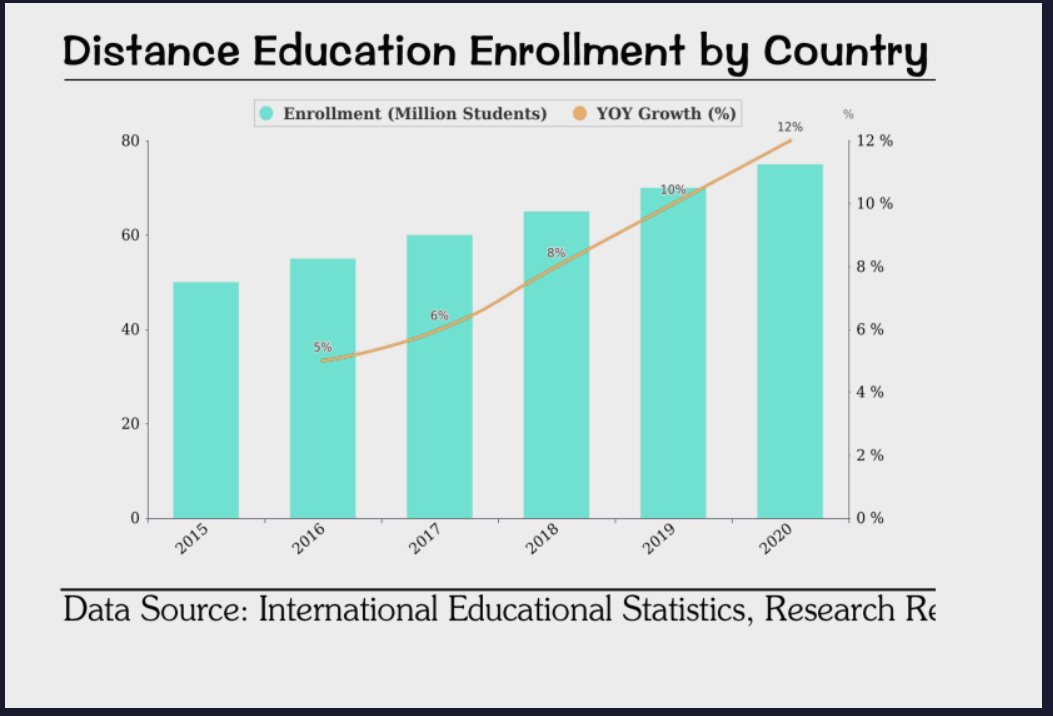} \\
        \hline


        \makecell[l]{Text\\Grounding} &
        Where is the region of the text "HOLLYWOOD"? Output the normalized coordinates of the left-top and right-bottom corners of the bounding box. The coordinates should be normalized ranging from 0 to 1000 by the image width and height.\newline
        Your answer should be in the following format:\newline
        (x1, y1, x2, y2) \# x1, y1, x2, y2 are the normalized coordinates of the bounding box. &
        (209, 292, 411, 335) &
        \tableimage{} \\
        \hline

        \makecell[l]{Key Information\\Mapping} &
        According to the information in the image, please pair the corresponding keys and values below: Keys that need to be paired are `Total Fat per 100g/ml', `Protein per 100g/ml'. Values that need to be paired are `5.0g', `1.4g'. Extract the required information in the text, and return the result in a dict directly. Your answer should be in the JSON format:\newline
        \{``key1'': ``..'', ``key2'': ``..'', \ldots\} &
        \{``Total Fat per 100g/ml'': [``1.4g''], ``Protein per 100g/ml'': [``5.0g'']\} &
        \tableimage{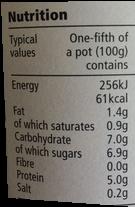} \\
        \hline

        \makecell[l]{Handwritten\\Answer\\Extraction} &
        在单项选择题第4题中，考生答题的手写内容是什么？请输出考生选择的选项 &
        B &
        \tableimage{} \\
        \hline

        \makecell[l]{Document\\Classification} &
        What type of document is the image? Here is the options: letter, form, email, handwritten, advertisement, scientific report, scientific publication, specification, file folder, news article, budget, invoice, presentation, questionnaire, resume, memo. Directly output the results without any additional explanation. &
        memo &
        \tableimage{} \\
        \hline

        Diagram QA &
        What does the rabbit eat? Snake, Grass, Hawk, or Frog? &
        Grass &
        \tableimage{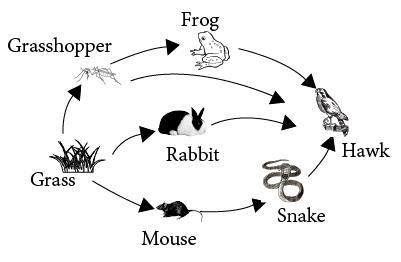} \\
        \hline

        Math QA &
        Please use the information from the provided image to directly answer the mathematical question described in the image. &
        8 &
        \tableimage{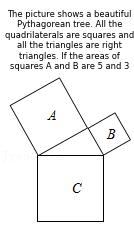} \\
        \hline

        Reasoning VQA &
        what was the total number of points for alain prost and nelson piquet? &
        15 &
        \tableimage{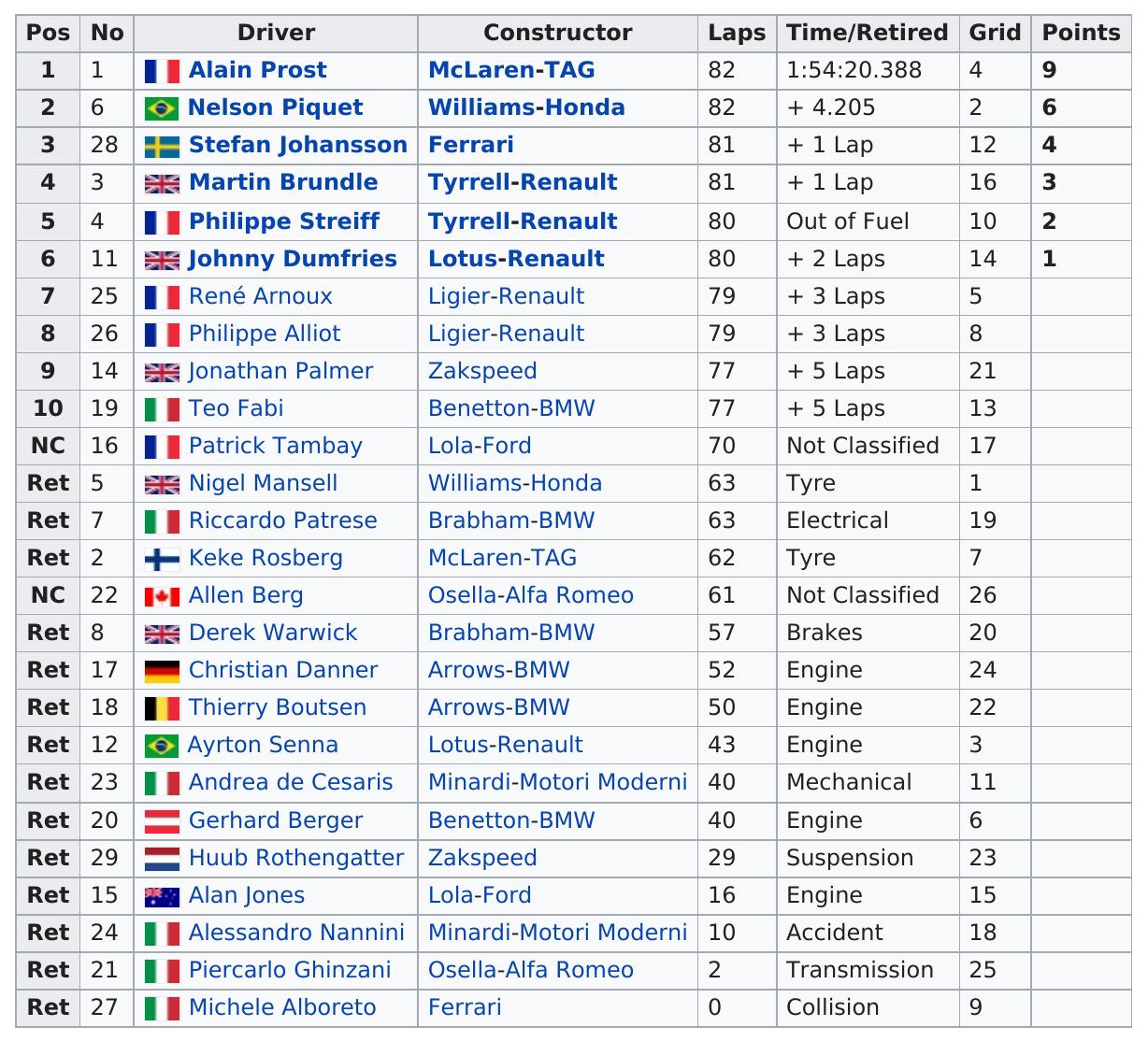} \\
        \hline

\end{longtable}

\endgroup
\end{CJK}

\end{document}